\documentclass{article} % For LaTeX2e
\usepackage{iclr2025_conference,times}

\usepackage{amsmath,amsfonts,bm}

\def\eqref#1{equation~\ref{#1}}
\def\1{\bm{1}}

\DeclareMathAlphabet{\mathsfit}{\encodingdefault}{\sfdefault}{m}{sl}
\SetMathAlphabet{\mathsfit}{bold}{\encodingdefault}{\sfdefault}{bx}{n}

\newcommand{\lr}{\alpha}

\DeclareMathOperator*{\argmax}{arg\,max}

\usepackage{hyperref}
\usepackage{url}

\usepackage[most]{tcolorbox}
\tcbuselibrary{skins}  % for tcbline
\usepackage{float}
\usepackage{cleveref}  % for cref
\usepackage{amssymb}  % for mathbb
\usepackage{bm}  % for bm
\usepackage{algorithm}  % for pseudocode
\usepackage{algpseudocode}  % for pseudocode
\usepackage{multirow}   % for table with multirow
\usepackage{adjustbox}  % to use adjustbox
\usepackage{subcaption}
\usepackage{colortbl}
\usepackage{graphicx}
\usepackage{wrapfig}  % to make the text surround table
\usepackage{booktabs}
\usepackage{makecell}

\newfloat{figtab}{htb}{fgtb}
\makeatletter
  \newcommand\figcaption{\def\@captype{figure}\caption}
  \newcommand\tabcaption{\def\@captype{table}\caption}
\makeatother

\newcommand{\St}{{\mathcal S}}
\newcommand{\Ac}{{\mathcal A}}
\newcommand{\Trans}{{\pw{P}}}

\newcommand{\Q}{{Q}} % Q function, which is a real number
\newcommand{\Expect}{{\mathbb E}}
\newcommand{\Loss}{{\mathcal L}}
\newcommand{\LossOneTD}{{\Loss_{1}^{RL}}}  % here I use 'RL' instead of 'TD' since I didn't mention the concept of 'TD' in the preliminary section.
\newcommand{\aE}{a^L}

\newcommand{\LossNTD}{{\Loss_{\StepN}^{RL}}}

\newcommand{\LossMarg}{{\Loss^{MG}}}
\newcommand{\LossMargETGT}{{\Loss^{MG}_{L+TGT}}}
\newcommand{\LossMargE}{{\Loss_{L}^{MG}}}

\newcommand{\LossMargTGT}{{\Loss_{TGT}^{MG}}}
\newcommand{\WLossMargTGT}{{\bar{w}}}
\newcommand{\MarginConst}{C}
\newcommand{\LossPVP}{{\Loss_{L}^{PVP}}}

\newcommand{\LossOneTDwoR}{{\Loss_{1(r=0)}^{RL}}}

\newcommand{\LossNTDwoR}{{\Loss_{\StepN(r=0)}^{RL}}}

\newcommand{\LossPropETGT}{{\mathcal{L}^{Prop}}}

\newcommand{\ours}{{ICoPro}} % name of our model/method
\newcommand{\oursHuman}{{ICoPro-Human}} % name of our model/method
\newcommand{\Rainbow}{{Rainbow}} % baseline: data-efficient Rainbow
\newcommand{\BC}{{BC-L}} % baseline: behavior cloning
\newcommand{\DQfD}{{DQfD-I}} % baseline: DQfD in our iterative learning scheme
\newcommand{\PVPwoR}{{PVP-I}} % baseline: PVP in our iterative learning scheme without reward (recommended setting in the PVP paper)
\newcommand{\PVPwR}{{PVP-IR}} % baseline: PVP in our iterative learning scheme with reward
\newcommand{\AblaNoFT}{{w/o $\PhaseOne$}} % ablation for ours
\newcommand{\AblaNoTGT}{{w/o $a^{TGT}$}} % ablation for ours
\newcommand{\DAgger}{{HGDAgger-L}} % baseline & ablation for ours (noFinetune, no TGT)

\newcommand{\phaseData}{{\tt data collection}} % name of our model/method
\newcommand{\PhaseData}{{\tt Data Collection}} % name of our model/method
\newcommand{\PhaseDataTitle}{{Data Collection}} % name of our model/method

\newcommand{\phaseOne}{{\tt finetune}} % name of our model/method
\newcommand{\PhaseOne}{{\tt Finetune}} % name of our model/method
\newcommand{\PhaseOneTitle}{Finetune} % name of our model/method

\newcommand{\phaseTwo}{{\tt propagation}} % name of our model/method
\newcommand{\PhaseTwo}{{\tt Propagation}} % name of our model/method
\newcommand{\PhaseTwoTitle}{{Propagation}} % name of our model/method

\newcommand{\LenRollout}{H} % length of the whole trajectory
\newcommand{\LenSeg}{T} % length of the query segment
\newcommand{\Query}{q}
\newcommand{\CntQueryItr}{N_q}  % the current iteration
\newcommand{\CntCFSeg}{N_{CF}}
\newcommand{\Qa}{Q^{\theta}} % Q function of the training agent
\newcommand{\QE}{{Q^{L}}} % Q function of the expert
\newcommand{\QTarget}{{\bar\Q}} % target Q
\newcommand{\pia}{\pi^{\theta}} % Q function of the training agent
\newcommand{\piE}{{\pi^{L}}} % Q function of the expert
\newcommand{\StepN}{n} % length of the segment
\newcommand{\EpsG}{\mathcal{G}_{\epsilon}}  % epsilon-greedy

\newcommand{\EnvData}{\mathcal{D}^{env}} % (use superscript since the subscript is iteration) data from interacting with the env
\newcommand{\LabelDataE}{\mathcal{D}^{L}} % label buffer of CFs
\newcommand{\LabelDataTgt}{\mathcal{D}^{TGT}} % label buffer of CFs
\newcommand{\LenLabelData}{N_{lab}} % label buffer of CFs
\newcommand{\TotalIter}{N_{Itr}}  % the total number of iterations in our method
\newcommand{\CurrentIter}{i}  % the current iteration
\newcommand{\AccTarget}{\delta_{acc}}  % the current iteration
\newcommand{\RLIter}{I}  % the current iteration
\newcommand{\RLEpo}{E}  % the current iteration

\newcommand{\bs}{B}  % batch size

\newcommand{\proxyr}{\widetilde{r}}

\newcommand{\CrashRate}{{\%Crash}}
\newcommand{\StepAvg}{{Step-Avg}}
\newcommand{\StepMin}{{Step-Min}}
\newcommand{\DistanceAvg}{{Distance-Avg}}
\newcommand{\DistanceMin}{{Distance-Min}}
\newcommand{\SpeedAvg}{{Speed-Avg}}
\newcommand{\LanePosAvg}{{LanePosition-Avg}}
\newcommand{\LaneChangeAvg}{{\%LaneChange-Avg}}
\newcommand{\ExpertCL}{{Labeler-CL}}  % the expert that prefers to change lane when driving
\newcommand{\ExpertRL}{{Labeler-RL}}  % the expert that prefers to driving in the right lane
\newcommand{\DiffCkpt}{{\texttt{DiffLabler}}}
\newcommand{\DiffRand}{{\texttt{DiffRand}}}

\newcommand{\seaquest}{{Seaquest}}
\newcommand{\boxing}{{Boxing}}
\newcommand{\battlezone}{{Battlezone}}
\newcommand{\frostbite}{{Frostbite}}
\newcommand{\alien}{{Alien}}
\newcommand{\hero}{{Hero}}
\newcommand{\mspacman}{{MsPacman}}
\newcommand{\pong}{{Pong}}
\newcommand{\freeway}{{Freeway}}
\newcommand{\ed}{{Enduro}}
\newcommand{\proxyrAtari}{{\proxyr_{A}}}
\newcommand{\rawrAtari}{{r_{A}}}

\newtoggle{final}
\toggletrue{final}
\newcommand{\pw}[1]{\iftoggle{final}{#1}{{\color{blue} #1}}}
\newcommand{\zj}[1]{\iftoggle{final}{#1}{{\color{magenta} #1}}}

\newcommand{\xf}[1]{\iftoggle{final}{#1}{{\color{green} #1}}}

\newcommand{\CMT}[1]{\iftoggle{final}{}{\todo{#1}}}

\iftoggle{final}{}{\setlength{\marginparwidth}{3cm}}

\usepackage[textsize=tiny]{todonotes}
\title{Reinforcement Learning from Imperfect Corrective Actions and Proxy Rewards}

\author{Zhaohui Jiang, Xuening Feng \& Yifei Zhu\\
The University of Michigan-Shanghai Jiao Tong University Joint Institute (UM-SJTU JI)\\
\texttt{\{jiangzhaohui\}@sjtu.edu.cn} \\
\And
Paul Weng\\
Electrical and Computer Engineering, Duke Kunshan University \\
\texttt{\{paul.weng\}@dukekunshan.edu.cn} \\
\AND
Yan Song, Tianze Zhou, Yujing Hu, Tangjie Lv \& Changjie Fan\\
NetEase Fuxi AI Lab\\
}

\iclrfinalcopy % Uncomment for camera-ready version, but NOT for submission.
\begin{document}

\maketitle

\begin{abstract}
In practice, reinforcement learning (RL) agents are often trained with a possibly imperfect proxy reward function, which may lead to a human-agent alignment issue 
% (i.e., the agent learns a policy that exploits the imperfections in the definition of this reward function, instead of \zj{solving the task with expected performance style}).
\zj{(i.e., the \xf{learned} policy either converges to non-optimal performance with low cumulative reward\xf{s}, or achieves high cumulative rewards but in \pw{undesired} manner).}
%\CMT{Note that in the first case you talk about the true rewards, but in the second case you talk about the proxy rewards.
%The two cases are not exclusive and will probably happen at the same time.}
To tackle this issue, we consider a framework where a human \zj{labeler} can provide additional feedback \zj{in} the form of corrective actions,
% \xf{, which can express the labeler's preferences over actions although may not be globally perfect.}
\zj{which expresses the labeler's action preferences although \pw{this feedback may possibly be} imperfect as well}.
In this setting, \zj{to obtain a better-aligned policy \xf{guided by }% combining 
both learning signals}, we propose a novel value-based deep RL algorithm called \textbf{I}terative learning from \textbf{Co}rrective action\pw{s} and \textbf{Pro}xy reward\pw{s} (ICoPro)\footnote{We have open-sourced its implementation: \href{https://github.com/JiangZhaoh/RLHF_CF}{https://github.com/JiangZhaoh/RLHF\_CF}.}, which \zj{cycles through three} phases: 
\zj{(1) Solicit sparse corrective actions from a human labeler on the agent's demonstrated trajectories; 
(2) Incorporate these corrective actions into the Q-function using a margin loss to enforce adherence to labeler's preferences; 
(3)} Train the agent with standard RL loss\zj{es} regularized with \xf{a} margin loss to learn from proxy rewards and propagate the Q-values learned from human feedback.
\zj{Moreover, another novel design} in our approach is to integrate pseudo-labels from the target Q-network to reduce human labor and further stabilize training. 
We \pw{experimentally} validate our proposition on a variety of tasks (Atari games and autonomous driving on highway\zj{).} 
\pw{On the one hand, u}sing proxy rewards with different levels of imperfection, our method can better align with \xf{human} \pw{preferences} \zj{and is more sample-efficient than baseline methods.
\pw{On the other hand, f}acing corrective actions with different types of imperfection, our method can overcome \pw{the} non-optimality \pw{of this feedback} thanks to \pw{the} guidance from proxy rewards.}
%\Note{xf: I think the last two sentences are duplicated. "better align with the labeler" implies that our method can overcome the non-optimality issue, no?}
\end{abstract}

\section{Introduction}

While reinforcement learning (RL) has proved its effectiveness in numerous application domains \citep{mnih2015DQN,silver2017AlphaGo,levine2016endVisuomotor},
% However, RL generally requires a huge amount of data (e.g., millions of transitions in Atari games) before the RL agent can learn a good behavior, making online RL training impractical in many domains.
% In addition, 
its impressive achievements are only possible if a high-quality reward signal for the RL agent to learn from is available. 
In practice, correctly defining such reward signal is often very difficult (e.g., in autonomous driving).
If rewards are misspecified, the RL agent would generally learn behaviors that are unexpected \citep{amodei2016concretePAISafety} and unwanted \citep{clark2016faultyReward,russell2016artificial} by the system designer, notably due to overoptimization \citep{gao2023ScalingLawRewardLearning} or specification gaming \citep{RandlovAlstrom98}.
This important issue has been well-recognized in the research community \citep{AmodeiOlahSteinhardtChristianoSchulmanMane16} and is an active research direction \citep{ouyang2022InstructGPT,skalse2022definRewardHack}.

Various solutions have been proposed to avoid having to define a reward function, for instance, 
behavior cloning \citep{Pomerleau89,bain1995frameworkBC}, inverse reinforcement learning \citep{ng2000IRL,Russell98}, (inverse) reward design \citep{NIPS2017InverseRewardDesign}, or reinforcement learning from human feedback \pw{\citep{BusaFeketeSzorenyiWengChengHullermeier14,christiano2017RLHF}}.
% \CMT{Actually, we worked on this kind of topic very early.}
However, these approaches could be impractical and inefficient \zj{since} they may require a perfect demonstrator, a reliable expert to provide correct labels, or assume that the agent only learns from human feedback, which would subsequently require too many (often hard-to-answer) queries for a human to consider.

As an alternative intermediate approach, we propose the following framework in which the RL agent has access to two sources of signals to learn from: proxy rewards and corrective actions.
A \textit{proxy reward} function is an imperfect reward function, that approximately specifies the task to be learned.
A \textit{corrective action} is provided by a (possibly unreliable) human \zj{labeler} when s/he is queried by the RL agent: 
a trajectory segment is shown to the labeler, who can then choose to correct an action performed by the agent by demonstrating another supposedly-good action.

\pw{This} framework is practical and easily \pw{implementable}.
Regarding proxy rewards, they are generally easy for system designers to provide.
For instance, they can 
(1) learn proxy rewards from (possibly imperfect) demonstration, 
(2) manually specify them to roughly express their intention (e.g., supposedly-good actions are rewarded and expected bad actions are penalized), or 
(3) only define sparse rewards (e.g., positive reward for reaching a destination and penalty for a crash).
Regarding corrective actions, in contrast to typical demonstrations of whole trajectories, \zj{this feedback is usually much easier for the labeler to provide, since humans may not be able to complete a task themselves but can readily offer action preference on some states.}
%, this feedback is usually much easier to provide since only one action is required.

\pw{While only learning with one of those two sources of signals has its own limitations, by proposing our framework, we argue (and experimentally demonstrate) that} 
learning \pw{simultaneously} from \pw{both of them}, even though both may be imperfect, can be highly beneficial.
\pw{More specifically, on the one hand}, solely learning from proxy rewards would either lead to very slow learning (e.g., when proxy rewards are well-aligned with ground-truth rewards but are very sparse) or yield a policy whose performance is not acceptable to the system designer (e.g., when proxy rewards are dense, but misspecified). 
\pw{On the other hand}, solely learning from corrective actions would require too many queries to the human labeler and may lead to suboptimal learned behaviors since the human may be suboptimal.
\pw{In contrast}, our key insight is that the two sources of signals can complement each other.
Since they are generally imperfect in different state-space regions, bad decisions learned from proxy rewards can be corrected by the human \zj{labeler}, while the effects of suboptimal corrective actions may be weakened by proxy rewards.
Therefore, learning simultaneously from the two imperfect signals can achieve better behaviors more aligned to the system designer's goals than using any one of the two alone and can be more data efficient (in terms of both \zj{environmental} transitions and \zj{human} queries).

\begin{figure}
  \centering
  \includegraphics[width=0.9\textwidth]{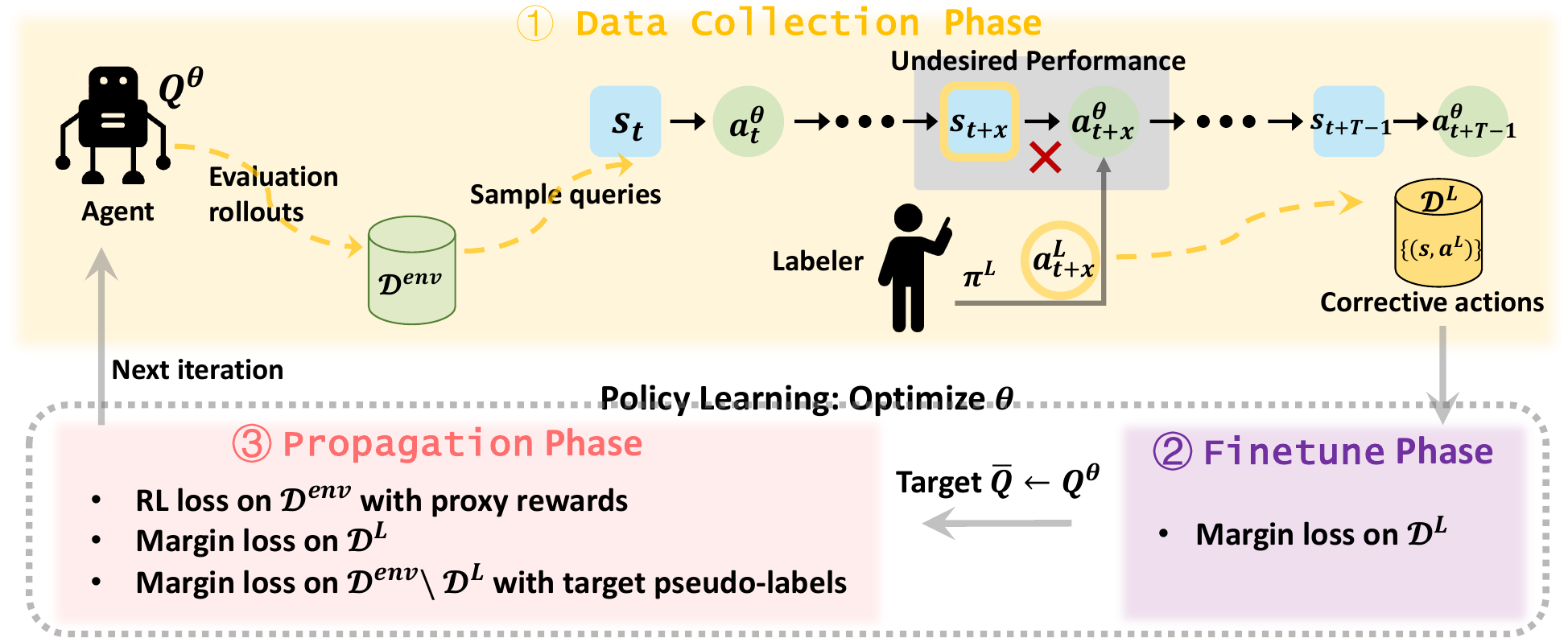}
  % \vspace{-0.5cm}
  \caption{\zj{\ours{} is an iterative method with \zj{three phases} in each iteration. It starts with the \phaseData-phase to collect agent's \zj{rollouts}. Segments are then sampled from these rollouts and used as queries for the labeler to provide several corrective actions. Following this are two separate phases for policy updating, the \phaseOne- and \phaseTwo-phase. Then the updated policy is utilized in the \phaseData-phase of the next iteration.}}
    \label{fig:FrameworkIllustration}
\end{figure}

As a proof of concept, we design a novel \zj{value-based} RL algorithm (see \Cref{fig:FrameworkIllustration}), called Iterative learning from Corrective action and Proxy reward (\ours{}).
% \zj{which is instantiated as a value-based method using the well-known model architecture of Rainbow \citep{hessel2018Rainbow}.}
% , using as base algorithm, Rainbow \citep{hessel2018Rainbow}, a state-of-the-art value-based deep RL algorithm. 
% \CMT{zj: I think it may be too early to mention Rainbow here: ``using as base algorithm, Rainbow \citep{hessel2018Rainbow}, a state-of-the-art value-based deep RL algorithm''.  And we have mentioned it again at the beginning in \Cref{sec:method}.}
\ours{} alternates between \zj{three} steps.
In the first \zj{step} ($\phaseData$\zj{-phase}), 
% \CMT{The other ones has "phase" in the name, but not this one. 
% Is it normal?  zj: add phase here is better}
the RL agent interacts with the environment to collect transition data, and then the labeler provides corrective actions on them.
In the \zj{second} step ($\phaseOne$-phase), the RL agent learns to select actions according to this feedback via a margin loss, which can be interpreted as an imitation learning \zj{(IL)} loss.
In the \zj{third} step ($\phaseTwo$-phase), the RL agent is trained to maximize the expected cumulative proxy rewards while further enforcing a margin loss.
This latter loss is expressed on both observed labels (i.e., corrective actions given by the human \zj{labeler}) and pseudo-labels generated by a trained model (target Q-network).
Imitating \zj{such} pseudo-labels can be interpreted in two ways: either to reduce the number of queries or to stay close to the previously-learned policy, which can stabilize training like in trust region methods \citep{SchulmanLevineAbbeelJordanMoritz15,SchulmanWolskiDhariwalRadfordKlimov17,asadi2022DQPro}.
By combining imitation learning and reinforcement learning in this third phase, the agent can learn from both proxy rewards and enforce the temporal consistency of the values learned from imitation learning.

The contributions of this paper can be summarized as follows:
(1) We propose a practical human-in-the-loop reinforcement learning framework (\Cref{sec:ProblemFormulation}) where the RL agent can learn from proxy rewards and (possibly suboptimal) corrective actions.
(2) To train the agent from these two imperfect sources, we design an efficient and robust algorithm (\Cref{sec:method}) in which we demonstrate a simple but useful technique to reduce human feedback. 
(3) We experimentally validate our proof of concept (\Cref{sec:experiments}) by conducting an extensive number of experiments on various domains (e.g., \zj{image-based} Atari games and \zj{state-based} highway driving with sparse and dense proxy rewards) under different conditions (e.g., \zj{simulated or real human labeler}).

\section{Related work}
\label{sec:related}

Given the difficulties in defining ground-truth rewards, a growing number of studies attempt to incorporate various types of human feedback \zj{to train agents}, for example, 
    demonstrations \citep{Pomerleau89,hester2018DQfD,reddy2019SQIL},
    preferences \citep{BusaFeketeSzorenyiWengChengHullermeier14,christiano2017RLHF,lee2021pebble},
    scalar evaluation \citep{knox2012TAMER,knox2012RLTAMER,saunders2018TrialWoErr}, % TAMER: (TODO: cite TAMER, TAMER+RL) TAMER is a framework to map numerical human feedback to a reward function and utilize the learned reward model to guide RL training and/or the behaviour policy.
    action advising \citep{MaclinShavlik96,da2020uncertaintyAgentAdvise,ilhan2021AAAN},
    interventions \citep{peng2023PVP,luo2024RLIF},  % usually human take over control for a while
    example-states \citep{eysenbach2021ExampleStates},
    or \pw{combination of} multiple feedback types \citep{ibarz2018PbDQfD,jeon2020UnifyRewardLearning,yuan2024uniRLHF,dong2024AlignDiff}.
Among them, demonstrations, action advising, and interventions, which we discuss further below, are the most related to corrective actions.
We roughly organize these works in three clusters.
Preference signals can be provided in an \emph{offline} or \emph{online setting}.
In addition, we discuss related work that considers \emph{combining reward and other preference signals}.

\paragraph{Offline setting}
One of the earliest approaches to circumvent the need of defining a reward function is 
imitation learning (\pw{aka} learning from demonstration)\pw{.} %\citep{bain1995frameworkBC,Pomerleau89}.
In this setting where a batch of %\zj{labeler} 
%%% Not sure why it's useful to specify labeler. In the offline, labeler is not a common term, usually: demonstrator or expert is used.
demonstration is assumed to be available, two main approaches have been developed:
behavior cloning \citep{Pomerleau89,bain1995frameworkBC} and inverse reinforcement learning \citep{ng2000IRL,Russell98}, with recent instantiations \citep{ho2016GAIL,reddy2019SQIL} trying to overcome the limitations of these two approaches.
%In the former, the agent directly tries to mimic the decisions observed in the demonstration, while in the latter, a reward function is first recovered from demonstration, before being used to learn a policy.
Most of these methods require  demonstrations \pw{to be} (near) optimal.
Since a batch of demonstration\zj{s} is provided, they may also suffer from train-test distribution mismatch.
A typical approach to tackle this latter issue is to resort to the online setting.

\paragraph{Online setting}
In contrast to the \pw{offline} setting, here the agent learns with a human in the loop\zj{:}
either a human actively oversees the agent's training and can intervene, or the agent can query a human to obtain feedback to learn from. 
The former case \citep{spencer2020EIL,peng2023PVP,luo2024RLIF} requires active participation and continuous attention \zj{from} humans, while the latter may require too many queries, %\zj{particularly if there is only one source} of preference data, 
which can be \pw{too} \zj{labor \pw{intensive} for \pw{one or even multiple humans} to provide}.
Examples of the latter approach include \pw{methods in} \zj{the DAgger \pw{family}} \citep{Ross2011DAgger,sun2017DeepAggreVaTeD,Kelly2019HGDAgger} \pw{(with optimal action labeling)}, and reinforcement learning from human feedback (RLHF) methods \citep{BusaFeketeSzorenyiWengChengHullermeier14,christiano2017RLHF,lee2021pebble} \pw{(with pairwise comparisons)}.

\paragraph{Combining reward and other signals}
Since learning from one signal may be insufficient and/or inefficient, multiple works \citep{hester2018DQfD,ibarz2018PbDQfD} consider learning from demonstration and rewards.
In such work, demonstration is often provided offline and used for initializing an online training phase, which can be standard RL training \citep{hester2018DQfD} or use other types of feedback such as in RLHF \citep{ibarz2018PbDQfD}, which can notably help for better human-agent alignment.
In particular, demonstrations can be used to initialize a policy \citep{ibarz2018PbDQfD},  define a reward signal \citep{reddy2019SQIL}, or \pw{initialize} a replay buffer for future RL training \citep{hester2018DQfD}.
In addition, they can also be queried online \citep{chen2020ActiveDQN,peng2023PVP,luo2024RLIF}.
% Like in the offline setting, complementing demonstration with  \zj{labeler} actions collected the agent's trajectories have been investigated \citep{EIL}.
\zj{Most works assume \pw{a well-specified (ground-truth)} reward function, then focus on balancing the IL and RL objectives 
% \CMT{Has IL been defined before as an acronym? zj: previously no, now I have added this acronym in the introduction section.}
when human demonstrations are also \pw{usually assumed to be} optimal \citep{shenfeld23aTGRL,liu2024RobustPolicyImprove}, although some proposition\pw{s} have been made to deal with suboptimal demonstration\pw{s} \citep{nair2018ddpgDemoQFilter}.}
% Most work assumes that the human is optimal, although some proposition has been made to deal with suboptimal demonstration, but assuming a ground-truth reward function \citep{nair2018ddpgDemoQFilter}.
% In this \zj{former} case, other works \citep{shenfeld23aTGRL,liu2024RobustPolicyImprove} focus on balancing the IL and RL objectives.

% In contrast to previous work, we do not assume that a ground-truth reward function is available, but only an approximation of it (proxy reward), which is easy to specify, is provided to the agent.
\zj{In contrast to previous work, we do not assume the availability of a ground-truth reward function but only an approximation of it (i.e., a proxy reward), which is easier to specify.}
Since learning from this imperfect signal is insufficient, we complement it with corrective actions.
Providing corrective actions is less cognitively-demanding for the human than generating whole demonstration\zj{s}, especially if the agent and the human do not operate in the same \pw{sensori-motor} space.
% Compared to action labeling like in DAgger \citep{Ross2011DAgger}, corrective actions are less costly since the human chooses which state s/he wants to give the feedback.
\zj{Compared to action labeling \citep{Ross2011DAgger,Kelly2019HGDAgger} or intervention \citep{peng2023PVP,luo2024RLIF}\pw{, which have to happen on} all unsatisfactory cases, our corrective action setting is} less costly since the human chooses \pw{on} which states s/he wants to give the feedback.
Finally, compared to \pw{most} previous work, we do not assume that the human is optimal.
The goal of our paper is to demonstrate as a proof of concept that the agent can learn more efficiently to reach a better performance from two \pw{imperfect} sources of signals than with any of the two alone.

% \TODO{Use somewhere?:
% On the one hand, pure imitation learning methods require a large number of human labels, whether it is under the offline setting or the online HiL setting.
% On the other hand, pure RL methods need a huge number of timesteps for interaction with the environment, which can be unacceptable in realistic situations.
% Therefore, combining the two kinds of methods is promising in reducing both the human labour cost and the environmental interaction timesteps.
% }

\section{Problem Formulation}\label{sec:ProblemFormulation}

We consider a reinforcement learning (RL) problem \citep{sutton2018RLIntro} where an RL agent repeatedly interacts with an environment: 
At a time step $t$, after observing an (observation) state $s_t \in \St$, it performs an action $a_t \in \Ac$, which yields an immediate reward $r_t \in \mathbb R$ and brings the agent to a new state 
% $s_{t+1} \in \St$
\zj{$s_{t+1}\sim\Trans(\cdot|s_t, a_t)$ where $\Trans$ is the environmental transition function}.
The goal of the RL agent is to learn, from interaction tuples $(s_t, a_t, r_t, s_{t+1})$, a policy $\pi^* : \St \to \Ac$ that maximizes the expected discounted sum of rewards.
Recall that this policy can be represented as the greedy policy with respect to a $Q$-function, $Q:\St \times \Ac \to \mathbb R$, $\pi^*(s) = \argmax_a Q(s, a)$.

In contrast to standard RL, we do not assume that a \pw{\emph{ground-truth}} reward function \pw{(i.e., providing a perfect description of} the task that the RL agent needs to learn\pw{) is available.
Instead, the agent only receives an approximation of it, called \emph{proxy reward} function and denoted $\proxyr$}.
Indeed, while a precise reward function capturing all aspects of a desired behavior is hard to define for \pw{a system designer}
%human \zj{labeler}s, 
% \CMT{It's strange to use labeler here, since a labeler is someone who provides labels.
% I think it may be better to use "system designer" in this paragraph. zj: yes, it's a mistake to use labeler here, the original word here is 'expert'}
providing a proxy reward \pw{function} to roughly express the 
\pw{system designer's}
%\zj{labeler}'s 
objectives is much easier to \pw{achieve} \citep{reddy2019SQIL,jeon2020UnifyRewardLearning,luo2024RLIF}.
However, since the proxy rewards are approximate, purely relying on learning from them may never achieve the desired performance.
% \todo{We should probably make the terminology consistent: labeler vs \zj{labeler}}
Therefore, we assume that the RL agent can also query a human labeler to obtain additional feedback under the form of \emph{corrective actions} \zj{(see \Cref{fig:FrameworkIllustration})}.
% This additional learning signal can be obtained by issuing a query to a human labeler (see \Cref{fig:FrameworkIllustration}).
% \CMT{Does this last sentence provide any more information wrt to the one before? zj: no, I merge the two sentences.}
Formally, a \textit{query} corresponds to a $\LenSeg$-step segment (i.e., sequence of \pw{state-action} tuples) $q=\left(s_t,a_t,\dots,s_{t+\LenSeg-1},a_{t+\LenSeg-1}\right)$, which can for instance be sampled from trajectories generated from the interaction of the RL agent with its environment.
    %the training agent's evaluation trajectories with $a=\argmax_{a\in\Ac} Q_{\theta}(s,\cdot)$.
Given such query, the human labeler may select one state $s_{t'}$ \pw{(such that $t \le\! t' \le\! t+T-1$)} where s/he provides a corrective action $\aE_{t'} = \piE(s_{t'})$, 
% \CMT{Same question as last time: Should we consider a stochastic policy here? A human labeler would label in a stochastic way and in your experiments, if I remember correctly, you consider various non-optimal scripted labelers and some of them are stochastic. zj: yes, the \DiffRand{} setting in \Cref{subsec:rand_actions}.}
% \CMT{I changed the notation to $t'$ for two reasons: $x$ was not explained and $x$ is not a great notation for an integer. zj: ok get it}
% \TODO{Should we consider a stochastic policy here? zj: I think no.}
where $\piE$ is the labeler's policy.
The corrective action is supposedly a better choice than the actually-performed one $a_{t'}$.
Moreover, the labeler may choose to provide no feedback if all the actions in the segment are good.

Note that we do not assume that the labeler is necessarily optimal. 
This means that it is not possible to solely rely on the corrective actions (without proxy rewards) to learn a policy to achieve the desired performance, which would furthermore be impractical in terms of label collection cost for the human labeler.
Instead, by combining the two imperfect sources of learning signals, the RL agent could possibly learn a policy 
\zj{with \pw{a better} performance \pw{and in a} more sample-efficient} \pw{way}
% (w.r.t. a ground truth reward function)  % zj: I comment this line since I think it conflicts with our experiments: we didn't use the ground-truth reward to measure the performance, in highway we use performance metric, in atari we use the goal-conditioned proxy rewards.
than using proxy rewards or corrective actions alone.
Intuitively, the imperfections of the two signals probably do not lie in the same state-action space regions and therefore the two signals could correct each other.
We confirm this intuition in our experiments (see \Cref{subsec:exp_one_ckpt,subsec:rand_actions}).

\section{Method}
\label{sec:method}

% As a proof of concept, in this paper, we focus on the case where the action space is finite and therefore build on the Rainbow algorithm \citep{hessel2018Rainbow}.
% According to the human-agent interaction scheme described above, we design \ours{}, an iterative approach to learn $\Qa$ with desired policy $\pia(s)=\argmax_{a\in\Ac}\Qa(s,\cdot)$.
\pw{As a first work demonstrating the benefit of learning from \zj{the} two imperfect sources of feedback, 
% \CMT{Is this first part true? zj: I think so. But to be more precise, I add a word ``the'' here to be more confident.}
we assume the action space $\Ac$ is finite for simplicity.
Building on the Rainbow algorithm \citep{hessel2018Rainbow}, we design \ours{}, an iterative approach to learn $\Qa$ with desired policy $\pia(s)=\argmax_{a\in\Ac}\Qa(s,\cdot)$.}
At each iteration $\CurrentIter$, \ours{} updates the current $\Q^{\theta_\CurrentIter}$ to obtain the next $\Q^{\theta_{\CurrentIter+1}}$ according to the three phases: $\PhaseData$, $\PhaseOne$, and $\PhaseTwo$, which we elaborate on next.
% (see \Cref{algo:ours} in the Appendix).
\zj{\Cref{fig:FrameworkIllustration} outlines the proposed framework and \Cref{algo:ours} is provided in \Cref{appendix:PseeudoCodes}.}
% \CMT{Should there be a reference to \Cref{fig:FrameworkIllustration}? zj: yes}

\paragraph{\PhaseDataTitle-phase.} 
Inspired by the popular RLHF setting \citep{christiano2017RLHF}, we adopt 
\pw{the following protocol for collecting two types of data: interaction tuples and corrective actions.
For the former, the agent acts in its environment according to the $\epsilon$-greedy policy with \pw{respect to} $\Q^{\theta_\CurrentIter}$, \pw{denoted \zj{$\EpsG(\Q^{\theta_\CurrentIter})$},} to collect a set 
$\EnvData_\CurrentIter$ of $\LenRollout$ interaction tuples, $(s_t, a_t, \proxyr_t, s_{t+1})$, which are also stored in the transition replay buffer $\EnvData$.
For the latter, the RL agent can regularly during RL training issue a batch of sampled queries from $\EnvData_\CurrentIter$ to the labeler to obtain a set $\LabelDataE_\CurrentIter$
of pairs of state and corrective action $(s,\zj{\aE})$, which are also stored in the \zj{ feedback replay buffer} $\LabelDataE$.}
%\CMT{Do we need notation $\EnvData_\CurrentIter$ in other parts? If not, I would suggest to remove it to reduce the number of notations. zj: although the answer is no, I prefer to keep this notation since $\EnvData_\CurrentIter$ means that queries are sampled from newly generated transitions.}
The collected data, $\EnvData$ and $\LabelDataE$, are used to train the RL agent in the next two phases.
%the following agent-labeler interaction scheme.
%Our RL agent can regularly issue a batch of queries to the labeler during RL training to obtain additional corrective actions.
%In addition, the agent interacts with the environment to obtain interaction tuples $(s_t, a_t, \zj{\proxyr}_t, s_{t+1})$.
%Using both the labeler's feedback and the interaction \pw{tuples}, the agent can update its policy.
As a side remark, for simplicity, querying the labeler is currently implemented in our experiments in a synchronous way (i.e., the agent waits for the labeler's feedback)\zj{, but} this process could
possibly 
also be asynchronous.

%For the next phases, we introduce the following notations.
%For replay buffer $\EnvData$ (resp. $\LabelDataE$) and a positive integer $k \le i$, $\EnvData[k]$ (resp. $\LabelDataE[k]$) denotes the union of the $k$ latest collected sets, i.e., at iteration $i$, $\EnvData[k] = \cup_{j=i-k+1}^{i} \EnvData_j$ (resp. $\LabelDataE[k] = \cup_{j=i-k+1}^{i} \LabelDataE_j$).
    
\paragraph{\PhaseOneTitle-phase.} 
This phase is a pure supervised \pw{training phase} to learn from all the \pw{labeler's} corrective actions collected so far in $\LabelDataE$.
Formally, $\Q^{\theta_\CurrentIter}$ is updated with the following margin loss $\LossMargE$ \citep{NIPS2013APID,piot2014MarginLoss,hester2018DQfD,ibarz2018PbDQfD}:
    \begin{equation}\label{equ:LossLabel}
        \LossMarg(\theta_\CurrentIter \mid \LabelDataE) = \Expect_{(s,\zj{\aE})\in\LabelDataE} \left[\max_{a\in\Ac}\left[\Q^{\theta_\CurrentIter}(s,a)+l(\zj{\aE},a)\right]-\Q^{\theta_\CurrentIter}(s,\zj{\aE})\right],
    \end{equation}
where $l(\zj{\aE},a)=0$ if $a=\zj{\aE}$, and a non-negative \textit{margin} value $\MarginConst$ otherwise.
This loss amounts to enforcing that the corrective actions' Q-values should not be smaller than those of any other actions.
As common practice, $\Q^{\theta_\CurrentIter}$ is updated via mini-batch stochastic gradient descent using \Cref{equ:LossLabel}.
Following \pw{classical} methods \citep{christiano2017RLHF} in RLHF, the updates end when the actions predicted by the updated $\Q^{\theta_\CurrentIter}$ reaches a pre-defined accuracy 
$\AccTarget$:
$\mathbb P_{s\sim\LabelDataE}\left[\zj{\aE}=\argmax_a \Q^{\theta_\CurrentIter}(s, a)\right]>\AccTarget \,.$
    % on $\LabelDataE_{[\CurrentIter,\CurrentIter]}$:
% \begin{equation}
%     \mathbb P_{s\sim\LabelDataE}\left[\zj{\aE}=\argmax_a \Q^{\theta_\CurrentIter}(s, a)\right]>\AccTarget \,.
% \end{equation}

% \begin{equation}
%     \Acc(\Qa_{\CurrentIter},\LabelDataE) = \Expect_{s\sim\LabelDataE}\left[\mathbb{I}\big[a^E=\argmax_a\Qa_{\CurrentIter}(s, \pw{a})\big]\right]>\AccTarget \,.
% \end{equation}

\paragraph{\PhaseTwoTitle-phase.} 
Since pure imitation learning requires a large number of human labels, 
we design the $\phaseTwo$-phase to include the training of the updated $\Q^{\theta_i}$ with a combination of RL losses and margin loss (from actual labels in $\LabelDataE$ but also pseudo-labels), which we explain next.

\textbf{- \emph{Training with RL Losses}} allows not only learning from proxy rewards, but also propagating the effect of human labels to more states (learned in the previous phase).
The initial target $\QTarget$ in this phase is the $\Q^{\theta_i}$ obtained at the end of the $\phaseOne$-phase.
These RL losses are composed of two commonly used terms, $\LossOneTD$ and $\LossNTD$, which are respectively defined as:
\begin{align}\label{equ:1StepTDLoss}
    &\LossOneTD(\theta_i \mid \EnvData,\QTarget) = 
        % \Expect_{(s_t,a_t,r_t,s_{t+1})\sim\EnvData}\left[
        \Expect_{\EnvData}\left[
            \left(
                \Q^{\theta_i}(s_t,a_t)-\left(\zj{\proxyr}_t+\gamma\max_{a'\in\Ac}\QTarget(s_{t+1},a')\right)
            \right)^2
        \right] \mbox{ and}\\
\label{equ:NStepTDLoss}
    &\LossNTD(\theta_i \mid \EnvData,\QTarget) = \Expect_{\EnvData}\left[
        \left(
            \Q^{\theta_i}(s_t,a_t)-\left(\sum_{k=0}^{\StepN-1}\gamma^k \zj{\proxyr}_{t+k}+\gamma^{\StepN}\max_{a'\in\Ac}\QTarget(s_{t+\StepN},a')\right)
        \right)^2
    \right],
\end{align}
where $\EnvData$ is the transition replay buffer, %$\Q^{\theta_i}$ is the online network at iteration $i$, 
and 
$\QTarget$ is the target network from a historical version of $\Q^{\theta_i}$. 
Note that in the context of standard RL, \Cref{equ:NStepTDLoss} is actually not theoretically-founded, since the goal is to learn a greedy policy.
However, in our context, it can help propagate faster the Q-values learned from the corrective actions.

\textbf{- \emph{Training with margin loss using \pw{both} actual and pseudo-labels}} allows to complement and correct the training with proxy rewards.
\pw{Training} with observed corrective actions in $\LabelDataE$ \pw{using loss $\LossMarg$} 
prevents the agent to forget about the actual labels.
\pw{In addition,} training with pseudo-labels can reduce the cost of collecting human labels by leveraging the large number of unlabeled \pw{states} in $\EnvData$. 
Pseudo-labels can be generated, using  predicted greedy actions from the target $\QTarget$, \pw{only on unlabeled states, since}
\pw{otherwise, it is better to enforce the actual label than a pseudo-label.}
% \TODO{Explain how $\bar Q$ is updated and $\Qa$? zj: I put them in the implementation details section}
\pw{Formally, t}he margin loss using pseudo-labels can be expressed as follows:
\begin{equation}
    \LossMargTGT(\theta_{\CurrentIter} \mid \LabelDataTgt) = \Expect_{(s, a^{TGT})\in\LabelDataTgt}\left[\max_{a\in\Ac}\left[\Q^{\theta_\CurrentIter}(s,a)+l(a^{TGT},a)\right]-\Q^{\theta_\CurrentIter}(s,a^{TGT})\right],
\end{equation}
where $\LabelDataTgt = \{(s, a^{TGT}) \mid s \in \EnvData\backslash\LabelDataE, a^{TGT}=\argmax_{a\in\Ac}\QTarget(s,\cdot)\}$
\pw{and by}
abuse of notations, $\EnvData\backslash\LabelDataE$ denotes the states that have not received any actual corrective actions $a^E$.
%For states that do have real \zj{labeler}'s corrective action $a^E$, we will not enforce $a^{TGT}$ on them.
\pw{Intuitively, as the training process goes on, we expect that} the quality of the pseudo-labels improves\pw{, enhancing further the benefit of enforcing $\LossMargTGT$.}
%\pw{In a different context,} \citet{asadi2022DQPro} \pw{regularize} the online network's parameters to stay close to the target network's to improve learning efficiency.
% Therefore 
%\pw{Loss} $\LossMargTGT$ can be understood from a perspective of introducing such similar stickiness for the Q-values, but is via the margin loss in the policy space instead of the parameter space. 
\pw{As a side note, $\LossMargTGT$ can also be understood as regularizing the online network's parameters to stay close to the target network, which has been shown to be beneficial in terms of learning efficiency \citep{asadi2022DQPro}.
In contrast to this previous work, we regularize via the margin loss in the policy space instead of the parameter space.}

To sum up, the total loss in this phase is the sum of the RL losses and the margin losses\footnote{In \Cref{equ:LossPropETGT} we give equal weight to \pw{the} IL and RL losses, but unequal weights could be provided if we expect one source to be more suboptimal than the other.}:
\begin{align}
    &\LossPropETGT = \LossOneTD(\theta_\CurrentIter \mid \EnvData,\QTarget) + \LossNTD(\theta_\CurrentIter \mid \EnvData,\QTarget) + \LossMargETGT(\theta_\CurrentIter \mid \EnvData, \LabelDataE) \mbox{, } \nonumber\\
    &\mbox{where }\LossMargETGT(\theta_\CurrentIter \mid \EnvData, \LabelDataE) =
        (1-\WLossMargTGT)\LossMargE(\theta_\CurrentIter \mid \LabelDataE) +
        \WLossMargTGT \cdot \LossMargTGT(\theta_{\CurrentIter} \mid \EnvData), \label{equ:LossPropETGT}
\end{align}
% \TODO{We should probably simplify further the notations, e.g., $\WLossMargTGT$}
and hyperparameter $\WLossMargTGT$ controls how much weight is put on pseudo-labels.
In our implementation,
    $\LossOneTD$, $\LossNTD$ and $\LossMargTGT$ works on a same minibatch from $\EnvData$ \zj{in each gradient step},
    while $\LossMargE$ works on another \zj{one} of the same size from $\LabelDataE$ to relieve the unbalanced issue of \pw{these} two datasets.
    % \TODO{How does this solve the "data unbalanced" issue? zj: for example, if in each gradient step, we calculate the gradient with 10 env samples with TD but only 1 \zj{labeler}'s label data, then this update can have bad prediction accuracy on $\LabelDataE$ since in each gradient step it can overfit to such single \zj{labeler}'s label data}

% There has  DQNPro\citet{asadi2022DQPro} is a paper that is very close to our design component about the target label.
% It's a paper that makes the online network stay close to the target network on the parameter space, while we enforce similar stickiness via the margin loss.
    
\subsection{Other implementation details}
% \paragraph{Updating schedule for $\Qa$ and $\QTarget$.}
\textbf{Updating schedule for $\Qa$ and $\QTarget$.}
% When optimize $\mathcal{L}^{Prop}$ (\Cref{equ:LossPropETGT}), in each gradient step,
%     we using the same mini-batch $\overline{\EnvData}\sim\EnvData$ to optimize both
%         $\LossOneTD(\theta_\CurrentIter \mid \EnvData,\QTarget)$, 
%         $\LossNTD(\theta_\CurrentIter \mid \EnvData,\QTarget)$, 
%         and $\LossMargTGT(\theta_{\CurrentIter}\mid \LabelDataTgt)$.
% We find that optimizing $\LossMargTGT$ on a different mini-batch with the two RL losses can make \ours{}'s training quite unstable and even damage the whole performance.
% which suggests that $\LossMargTGT$ can be understand from the regularization perspective as we
% In the \phaseTwo-phase, the initial target $\QTarget$ is the one at the end of the \phaseOne-phase.
Since $a^{TGT}$ will change every time we update the target $\QTarget$, we update $\QTarget$ with  low frequency to stabilize training.
Concretely, there are $\RLEpo$ epochs in one $\phaseTwo$-phase that \zj{each epoch optimize} $\mathcal{L}^{Prop}$ through $\EnvData$.
% \CMT{each? Is there a missing word?}
At the end of each epoch, $\QTarget$ will be updated to the online parameter $\Q^{\theta_\CurrentIter}$.
If $\RLEpo=1$, pseudo-labels in $\phaseTwo$-phase are generated form the policy at the end of $\phaseOne$-phase.
If $\RLEpo>1$, pseudo-labels are changing after each epoch and can come from the (potentially) improved $\Qa$ after optimizing $\mathcal{L}^{Prop}$.

% \paragraph{Optimization for \ours{}'s iterative scheme.} 
\textbf{Optimization for \ours{}'s iterative scheme.} 
% As for the optimizer, we use Adam \citep{diederik2014Adam} and use \zj{a} same one for the two \zj{learning} phases
As for the optimizer, we use use \zj{a} same \zj{optimizer through} the two \zj{learning} phases
% \CMT{There are three phases in ICoPro. zj: I add the word ``learning'' to be more precise.}
in \ours{}.
Since \ours{} uses an iterative learning scheme, we follow the research from \citet{asadi2024ResetOptRL} that reset the 1st and 2nd order moment inside the Adam optimizer at the beginning of each learning phase.
The $\phaseOne$- and $\phaseTwo$-phase use the same learning rate $\lr$, which remains unchanged throughout the training procedure.

\section{Experiments}
\label{sec:experiments}
% \CMT{zj: I think I add too many footnotes, is it ok??
% Paul: You should check if everything you specify is really important. 
% Usually, I believe that you give too many low-level details that makes the exposition hard to follow (e.g., footnote 3).}
We first explain our general experiment\pw{al} settings in \Cref{subsec:GeneralExpSetups}.
\zj{To compare \ours{} with baselines and \pw{analyze} its design components on a large scale of experiments, we use simulated labelers to demonstrate and \pw{analyze} \ours{}'s performance in \Cref{subsec:exp_one_ckpt,subsec:rand_actions}.
% \Cref{subsec:HighwayExperiment,subsec:AtariExperiment}.
Then in \Cref{subsec:userStudy}, we further verify the alignment performance of \ours{} using real humans' action feedback.}

\subsection{General experimental setups}\label{subsec:GeneralExpSetups}
% In this section, we mainly explain the baselines we use and how we simulate \zj{labeler}s to launch a large scale of experiments.
%\zj{Considering the realistic feasibility of human labour, we limit to at most 1\% of the environmental transitions labeled in our experiments.}
\pw{To ensure reasonable labeling effort, only} at most 1\% of the environmental transitions \pw{are} labeled in our experiments.
\pw{Hyper-parameter values are given in} \Cref{appendix:HyperParameters} unless otherwise specified.
Detailed evaluation configurations can be checked in \Cref{appendix:EvaluationDetails}.
% \CMT{zj: recap the content in Appendix.}
% \CMT{zj: seems these notations is not mentioned later: We use $\CntQueryItr$, $\LenRollout$, and $\TotalIter$ to represent the number of queries per iteration, the length of rollouts per iteration, and the total number of iterations when explaining the configuration for experiments.}

\textbf{Baselines.}
% To compare the performance of learning from only one of the signals online,
%     we use Rainbow \citep{hessel2018Rainbow} as the one that learns purely from the proxy reward,
%     DAgger \citep{Ross2011DAgger} with $\LossMargE$ as the one that learns purely from the \zj{labeler}'s online corrective actions.
% We also evaluate behaviour cloning (\BC{}) using $\LossMargE$ (\Cref{equ:LossLabel}) on all collected labels $\LabelDataE$ at the end of training of \ours{}.
\zj{To compare \pw{with} the performance of learning from only proxy rewards,
    we use Rainbow \citep{hessel2018Rainbow} as a baseline.
To compare \pw{with} the performance of learning from only corrective actions using $\LossMargE$ (\Cref{equ:LossLabel}),
    we evaluate \pw{behavior} cloning (\BC{}) on all collected labels $\LabelDataE$ at the end of training of \ours{} as the one in offline style,
    and adapt HG-DAgger \citep{Kelly2019HGDAgger} (\DAgger{}) by removing the $\phaseTwo$-phase from \ours{} as the one in online style.}
% \zj{To compare the performance of learning from only corrective actions in ,
\zj{As for baselines involving learning from both signals}, since no previous work considers the exact same learning scheme as explained in \Cref{sec:ProblemFormulation},
    we adapt two state-of-the-art methods that were initially designed on normal RL methods into our iterative learning setting \zj{with \pw{the} same labeling schedule and budget},
    DQfD \citep{hester2018DQfD} (\DQfD{}) and PVP \citep{peng2023PVP} (\PVPwoR{}),
    to compare the performance when learning from both corrective actions and proxy rewards.
Basically, the adaption is canceling the $\phaseOne$-phase in \ours{}
    and replacing their loss functions with \ours{}'s $\LossPropETGT$ (\Cref{equ:LossPropETGT}) in the $\phaseTwo$-phase.
\DQfD{} can be seen as an ablation of \ours{} by removing $\LossMargTGT$ \zj{from} $\LossPropETGT$.
\zj{\PVPwoR{}'s loss replaces $\LossMargETGT$ to $\LossPVP=\Expect_{(s,a,a^E)\sim\LabelDataE}\left[|\Qa(s,a^E)-1|^2+|\Qa(s,a)+1|^2\right]$, but using zero rewards in $\LossOneTD$ and $\LossNTD$.}
We also test \pw{in \PVPwoR{}, the effects of using} proxy rewards instead of zeros (\PVPwR{}).
% \zj{Except for \Rainbow{} and \BC{}, baselines receiving}
More details about \pw{the} baselines \zj{and} other ablations mentioned in \Cref{subsec:exp_one_ckpt} can be checked in \Cref{appendix:baselines}.
% \CMT{zj: More explanations about baselines in appendix.}

\textbf{Environments.}
Our experiments involve both the state-based highway \citep{highway-env} environment and imaged-based Atari \citep{aitchison2023Atari5} environments.

\begin{wraptable}{r}{6.8cm} % wraptable needs to be put in the front of text
    \centering
    \caption{Engineered proxy rewards on highway. 
    % The PR$X$ format in the first row means ProxyReward-Group$X$.
    We normalize the rewards into the range $[-1,1]$ in each step by min-max scaling if Normalization is not None, where the min (max) reward is the sum of negative (positive) event rewards.}
    \label{tab:HighwayReward1ChangeLane}
    \scalebox{0.75}{
        \setlength\tabcolsep{2.2pt} % smaller space between columns
        \begin{tabular}{ll|ccccccc}
            \toprule
            % \multirow{2}{*}{\multicolumn{2}{*}{Configurations}} & \multicolumn{5}{c}{Proxy Rewards}\\
           % \multirow{2}{*}{\multicolumn{2}{c|}{Configurations}} & \multicolumn{5}{c}{Proxy Rewards}\\
           \multicolumn{2}{c|}{\multirow{2}{*}{Configurations}} & \multicolumn{5}{c}{Proxy Rewards}\\
            & & PRExp & PR1 & PR2 & PR3 & PR4\\
           \midrule
            \multirow{4}{*}{\makecell[c]{Rewards\\ for \\Events}} & Change lane & 0.2 & 0 & 0.2 & 0 & 0\\
            % & Normalized lane index & 0  & 0 & 0 & 0 & 0\\
            & High speed & 1.5 & 2 & 0.8 & 0 & 0\\
            & Low speed & -0.5 & -1 & 0 & 0 & 0\\
            & Crash & -1.7 & -1 & -1 & -1 & -1\\
            \midrule
            \multicolumn{2}{c|}{Normalization} & [-1,1] & [-1,1] & [-1,1] & [-1,1] & None\\
            \midrule
            \multicolumn{2}{c|}{Dense(D) or Sparse(S)} & D & D & -- & D & S\\
            \bottomrule
        \end{tabular}
    }
\end{wraptable}
Highway is an environment that allows for flexible design of diverse proxy rewards and evaluation metrics, but is relatively simpler than Atari in terms of action dimension ($|\Ac|=5$) and environmental complexity.
The basic goal in this environment is to drive a car on a straight road 
with multiple lanes.
% within a limited time.
The final performance of the trained vehicle can be measured with different metrics, e.g.,
    the ratio of episodes with a crash (\CrashRate),
    the average total forward distance in a given time limit (\DistanceAvg),
    the average speed (\SpeedAvg),
    and the ratio of taking a change lane action (\LaneChangeAvg).
% We design representative proxy rewards in different levels of imperfection
% % % % \footnote{Since we restrict the speed of the controlled vehicle to either the high or low-speed range, PRExp and PR1 can be seen as dense rewards that receive non-zero rewards every time-step, and PR2 is a less dense one that driving in low speed receives zero rewards.
% % % With the normalization setting, PR3 can also be seen as a dense reward but with less performance preference.
% % PR4 is the most imperfect one that only penalises crashes.}
% \CMT{zj: move this footnote to appendix: ``Since we restrict the speed of the controlled vehicle to either the high or low-speed range, PRExp and PR1 can be seen as dense rewards that receive non-zero rewards every time-step, and PR2 is a less dense one that driving in low speed receives zero rewards.
% With the normalization setting, PR3 can also be seen as a dense reward but with less performance preference.
% PR4 is the most imperfect one that only penalises crashes.''}
% by associating different events with different rewards in each timestep as shown in \Cref{tab:HighwayReward1ChangeLane}.
% \CMT{Move the previous content into the appendix.}
\zj{We design representative proxy rewards in different levels of imperfection by associating events with different rewards as shown in \Cref{tab:HighwayReward1ChangeLane}.}
% , which covers both dense or sparse, goal-conditioned or performance-specified rewards. In the following content, PRExp refers to the least imperfect one among them, and PR1 to PR4 has increasing level}

In contrast, Atari provides more complex environments with larger action dimensions and longer episode lengths, although with less flexibility in designing proxy rewards and diverse performance metrics.
Since the dimension of action space can be seen as a measurement of hardness for one game, 
    we first select 6 representative games with the full action dimension ($|\Ac|=18$) across various categories suggested by \citet{aitchison2023Atari5}:
    (1) Combat: \battlezone{} and \seaquest{},
    (2) Sports: \boxing{},
    (3) Maze: \alien{},
    (4) Action: \frostbite{} and \hero{}.
We also include smaller-action-dimensional but classical games to complement our evaluation: MsPacman and Enduro ($|\Ac|=9$), Pong ($|\Ac|=6$), and Freeway ($|\Ac|=3$).
% We also use smaller-action-dimensional but classical games to complement our evaluation:
    % \mspacman{} and \ed{} with 9 actions, \pong{} with 6 actions, and \freeway{} with 3 actions.
In this setting, cumulative episodic raw rewards from ALE are used to measure agent performance, with signed raw rewards serving as proxy rewards\footnote{Raw rewards $\rawrAtari$ in Atari can be seen as goal-conditioned rewards, but not all desired performances (e.g., achieve goals in the right way) are rewarded in $\rawrAtari$. Further discussion is available in \Cref{appendix:AtariEnvDetails}.}.
% \CMT{zj: remember to explain $sign(\rawrAtari)$, with a equation that pos+1, neg-1.}

\textbf{Simulated \zj{labeler}.}
% We use Q-values trained from Rainbow to simulate human feedback as described in \Cref{assume:CFHumanModel}. 
We use Q-values trained \pw{with} Rainbow to simulate human feedback with $Q^{diff}(s,a,\zj{\aE})=\QE(s,\zj{\aE})-\QE(s,a)$, 
    where $\QE(s,a)=\Expect_{\Trans,\piE}[\sum_{t=0}^{\infty}\gamma^t r^{\zj{L}}(s,a)|s_0=s,a_0=a]$ is the \zj{labeler}'s Q-function, 
    and 
    % $a^E=\EpsG(\argmax\QE(s,\cdot))$ 
    $\aE=\EpsG(\QE(s,\cdot))$ 
    is the action issued from the \zj{labeler}, 
    % $a=\EpsG(\argmax\Qa(s,\cdot))$ 
    $a=\EpsG(\Qa(s,\cdot))$ 
    is the executed action from the training agent.
% \CMT{Have you introduced $\Trans$ before? zj: I add related explanation at the beginning of Preliminary section.}
% \CMT{Keep your notation consistent for $\EpsG$. 
% You didn't use argmax before. zj: ok,}
% \CMT{I believe that $a^E$ is not optimal. zj: I remove ``optimal'' when talking about $\aE$}
In practice, we select states with top $\CntCFSeg$ largest $Q^{diff}$ to give corrective actions for a querying segment \citet{luo2024RLIF}
% \footnote{A similar assumption based on such Q-difference has also been adopted in \citet{luo2024RLIF} but is to simulate human intervention feedback, which supports the rationality of this kind of simulated human model.}
, where $\CntCFSeg=1$ is our default configuration.
Basically, simulated \zj{labelers are agent checkpoints
    trained with engineered rewards by Rainbow \zj{but}
    % with relatively longer timesteps than \ours{}, but 
    do not necessarily converge to optimal performances.
For highway, \zj{labeler}s are trained with PRExp mentioned in \Cref{tab:HighwayReward1ChangeLane}.
For Atari, \zj{labeler}s are trained with the signed raw rewards.
% \CMT{Have you explained what is a signed raw reward? zj: I mention it at the last sentence of section 5.1 - Experiments paragraph.}
Concrete configurations to train \zj{labeler}s can be checked in \Cref{appendix:SimulatedExperts}.}

% \CMT{zj: put this sentence somewhere: In highway, one example is to train a labeler to drive at a suitable speed and prefers to change lanes frequently (\ExpertCL) by using the proxy reward shown in the PRExp column in \Cref{tab:HighwayReward1ChangeLane} with 330k timesteps.}

% Organize the experimental results using our conclusion in the abstract.
% \pw{On the one hand, u}sing proxy rewards with different levels of imperfection, our method can better align with \xf{human} \pw{preferences} \zj{and is more sample-efficient than baseline methods.
% \pw{On the other hand, f}acing corrective actions with different types of imperfection, our method can overcome \pw{the} non-optimality \pw{of this feedback} thanks to \pw{the} guidance from proxy rewards.}

\begin{figure}[t]
    \begin{subfigure}{\textwidth}
      \centering
      \includegraphics[width=\linewidth]{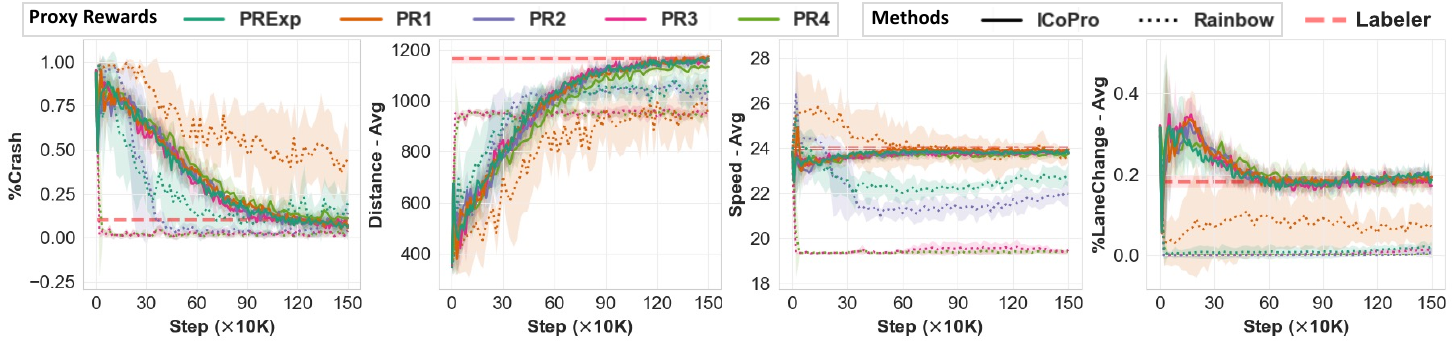}  
      \caption{Compare \ours{} \zj{(solid lines)} with \Rainbow{} \zj{(dotted lines)} using different proxy rewards \zj{(different colors)}.}
      \label{fig:HighwayC1vsGTLess}
    \end{subfigure}
    \newline
    \begin{subfigure}{\textwidth}
      \centering
      \includegraphics[width=\linewidth]{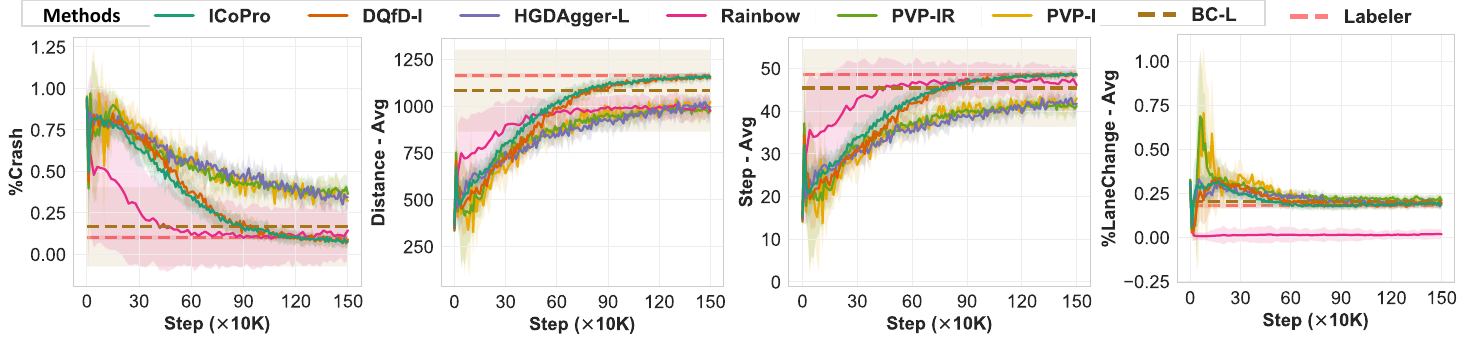}  
      \caption{Compare \ours{} with other baselines \zj{(different colors)}. Results are averaged over the set of proxy rewards.}
      \label{fig:HighwayC1BaselinesLess}
    \end{subfigure}
\caption{Experiments on highway over the set of proxy rewards. Subplots in one row compares the performance with respect to one performance metric. 
% Feedback schedule is $\CntQueryItr=10$, $\LenRollout=1k$, $\TotalIter=150$, and 
$|\LabelDataE|$=1.5K. % at the end of training.
% \zj{Dashed horizontal lines indicate the labeler's performance.}
}
\label{fig:highwayCkpt1ChangeLane}
\end{figure}

\subsection{Sample-efficient Alignment with Imperfect Proxy Rewards}\label{subsec:exp_one_ckpt}
We use one scripted labeler for each environment to validate the alignment ability of \ours{} and compare it with baselines, using proxy rewards with different levels of imperfections.
% Note that those scripted labelers are imperfect ones, since training longer with $r^L$ can lead to better performance.
In the following of this part, we \pw{analyze} \ours{}'s performance based on results shown in \Cref{fig:highwayCkpt1ChangeLane} and \Cref{tab:AtariBaselines}.
\pw{More specifically,} \Cref{fig:highwayCkpt1ChangeLane} evaluate\pw{s} the performance with various performance metrics, while \Cref{tab:AtariBaselines}  evaluate\pw{s} the performance with small or large size of action label budget\footnote{Considering that the action budgets required to align depend on the complexity of the labeler's performance, the concrete values of $|\LabelDataE|$ is not specified in the main text but can be check\pw{ed} in \Cref{tab:HypersDiffEnv}.}.
Due to space constraints, results for easier Atari tasks (i.e., \pong{}, \freeway{}, and \ed{}) and corresponding plots for \zj{the results in} \Cref{tab:AtariBaselines}, are put into \Cref{appendix:MoreExperiments}.
% \CMT{corresponding plots for a table?}
% Concretely, in highway, we use PRExp to obtain the \ExpertCL{} agent that can drive safely at decent speeds and prefers to change lanes. In Atari, 

% \textbf{Experiments and performance analysis.}

\begin{table}[t!]
\centering
\caption{Experimental results on Atari with different size of $|\LabelDataE|$. Red fonts denote the best results with the largest mean score. Bold black fonts emphasize results whose mean$+$std covers the best mean score. Yellow backgrounds highlight results that match the \zj{labeler}s' performance in the sense that their mean scores are in the range of the \zj{labeler}s' mean$\pm$std.}
\label{tab:AtariBaselines}
\scalebox{0.66}{% scale the table
\setlength\tabcolsep{2.6pt} % smaller space between columns
% \begin{tabular}{llccccccc}
\begin{tabular}{l|l|lllllll}
\toprule
\multirow{2}{*}{$|\LabelDataE|$} & \multirow{2}{*}{Methods} & \multicolumn{7}{c}{Environments}   \\
 &  & \seaquest{} & \boxing{} & \battlezone{} & \frostbite{} & \alien{} & \hero{} & \mspacman{}  \\
\midrule
 \multirow{2}{*}{$\backslash$}& \zj{Labeler} & 1071.92\footnotesize{$\pm$224.88} & 66.05\footnotesize{$\pm$9.17} & 24485.19\footnotesize{$\pm$4987.60} & 5647.63\footnotesize{$\pm$1038.91}  & 2908.97\footnotesize{$\pm$908.32}     & 26866.07\footnotesize{$\pm$612.29} & 3539.54\footnotesize{$\pm$1270.95}      
% & 33.00\footnotesize{$\pm$0.00}        & 21.00 \footnotesize{$\pm$0.00}      & 1253.36\footnotesize{$\pm$198.95}     
\\
 & \Rainbow{} & 707.04\footnotesize{$\pm$98.96}        & 1.99\footnotesize{$\pm$0.71}        & 17058.00\footnotesize{$\pm$2310.60}       & 2766.40\footnotesize{$\pm$518.78}        & 1319.83\footnotesize{$\pm$457.06}     & 19413.57\footnotesize{$\pm$600.39}        & 1773.98\footnotesize{$\pm$323.19}      
% & 32.16\footnotesize{$\pm$0.65}        & 18.40\footnotesize{$\pm$2.66}      & 759.50\footnotesize{$\pm$173.23}                  
\\
\midrule
\multirow{8}{*}{Large} & \DAgger{} & 790.10\footnotesize{$\pm$73.98} & 24.95\footnotesize{$\pm$2.22} & 16573.33\footnotesize{$\pm$1262.82} & 3207.35\footnotesize{$\pm$527.35} & 2067.45\footnotesize{$\pm$262.04}    & 13258.79\footnotesize{$\pm$2128.13}  & \cellcolor[HTML]{FFFFC7}\textbf{3052.38}\footnotesize{$\pm$370.50} \\
                                    & \PVPwoR{} & 635.83\footnotesize{$\pm$65.74}        & 7.76\footnotesize{$\pm$0.84}        & 15183.03\footnotesize{$\pm$1739.17}       & 1535.28\footnotesize{$\pm$528.88}        & 596.56\footnotesize{$\pm$99.45}      & 8503.46\footnotesize{$\pm$1205.13}      & 1686.13\footnotesize{$\pm$608.96} 
                                    %        & 26.58\footnotesize{$\pm$1.91}        & 0.28\footnotesize{$\pm$5.02}      & 255.51\footnotesize{$\pm$13.09}         
                                    \\
                                    & \PVPwR{}     & 770.88\footnotesize{$\pm$91.00}        & 13.36\footnotesize{$\pm$2.00}        & 17378.18\footnotesize{$\pm$2245.64}       & 1664.62\footnotesize{$\pm$414.02}        & 798.55\footnotesize{$\pm$137.99}     & 7729.12\footnotesize{$\pm$1282.99}       & 920.25\footnotesize{$\pm$216.76}       
                                    %  & 25.26\footnotesize{$\pm$0.88        & 20.40\footnotesize{$\pm$0.78      & 364.49\footnotesize{$\pm$13.11
                                    \\
                                     & \BC{}                                      & 758.40\footnotesize{$\pm$41.31}        & 21.39\footnotesize{$\pm$1.21}        & 16020.00\footnotesize{$\pm$2081.92}       & 3410.27\footnotesize{$\pm$479.08}        & \cellcolor[HTML]{FFFFC7}\textbf{\color{red}2360.40}\footnotesize{$\pm$422.57}     & 14288.67\footnotesize{$\pm$1673.48}       & \cellcolor[HTML]{FFFFC7}2877.07\footnotesize{$\pm$279.24}      
                                    % & 32.93\footnotesize{$\pm$0.03}        & 11.25\footnotesize{$\pm$6.16}      & 85.09\footnotesize{$\pm$8.00} 
                                    \\
                                    & \AblaNoTGT{}           & \cellcolor[HTML]{FFFFC7}\textbf{1213.98}\footnotesize{$\pm$102.54}       & 48.19\footnotesize{$\pm$7.90}        & 18111.52\footnotesize{$\pm$1409.68}       & 3874.36\footnotesize{$\pm$515.30}        & 1835.06\footnotesize{$\pm$267.85}     & 17805.35\footnotesize{$\pm$2471.35}       & \cellcolor[HTML]{FFFFC7}2579.86\footnotesize{$\pm$186.22} 
                                    % & 32.49\footnotesize{$\pm$0.46}        & 20.80\footnotesize{$\pm$0.19}      & 1130.53\footnotesize{$\pm$51.08}  
                                    \\
                                    & \AblaNoFT{}              & \cellcolor[HTML]{FFFFC7}\textbf{1234.39}\footnotesize{$\pm$113.44}       & 51.83\footnotesize{$\pm$26.67}       &           19366.67\footnotesize{$\pm$285.28}  & \cellcolor[HTML]{FFFFC7}\textbf{4713.07}\footnotesize{$\pm$1211.91}       & \cellcolor[HTML]{FFFFC7}\textbf{2223.66}\footnotesize{$\pm$368.38}     & 8398.37\footnotesize{$\pm$2729.85}       & \cellcolor[HTML]{FFFFC7}2633.59\footnotesize{$\pm$154.34}       
                                     %  &32.67\footnotesize{$\pm$0.24}        & 20.74\footnotesize{$\pm$0.29}      & 1027.75\footnotesize{$\pm$41.37}   
                                     \\
                                    & \DQfD{}                   & \cellcolor[HTML]{FFFFC7}1155.43\footnotesize{$\pm$85.98}        & 41.71\footnotesize{$\pm$5.09}        & 18444.85\footnotesize{$\pm$1279.70}       & 4176.27\footnotesize{$\pm$659.14}        & \cellcolor[HTML]{FFFFC7}2079.99\footnotesize{$\pm$293.53}     & 17814.10\footnotesize{$\pm$1688.55}       & \cellcolor[HTML]{FFFFC7}2734.25\footnotesize{$\pm$176.22} \\
                                    & \ours{}                    & \cellcolor[HTML]{FFFFC7}\textbf{\color{red}1274.76}\footnotesize{$\pm$109.53}       & \cellcolor[HTML]{FFFFC7}\textbf{\color{red}61.52}\footnotesize{$\pm$2.47}        & \cellcolor[HTML]{FFFFC7}\textbf{\color{red}20030.30}\footnotesize{$\pm$1620.36}       & \cellcolor[HTML]{FFFFC7}\textbf{\color{red}4817.74}\footnotesize{$\pm$605.45}        & \cellcolor[HTML]{FFFFC7}\textbf{\color{red}2360.27}\footnotesize{$\pm$301.52}     & \textbf{\color{red}23344.01}\footnotesize{$\pm$2353.67}       & \cellcolor[HTML]{FFFFC7}\textbf{\color{red}3188.20}\footnotesize{$\pm$453.13} \\
                                    % \cline{2-9}
                                    % & $|\LabelDataE|$ & & & & & & & \\
                                    \midrule
                                     
\multirow{8}{*}{Small}                  & \DAgger{}                   & 508.34\footnotesize{$\pm$77.91}        & 17.91\footnotesize{$\pm$2.21}        & 14682.22\footnotesize{$\pm$1633.46}       & 1561.04\footnotesize{$\pm$457.61}        & \textbf{1582.74}\footnotesize{$\pm$467.88}     & 10617.77\footnotesize{$\pm$1664.62}       & \cellcolor[HTML]{FFFFC7}\textbf{2342.51}\footnotesize{$\pm$264.97}      
% & 23.01\footnotesize{$\pm$0.78}        & -9.14\footnotesize{$\pm$5.09}      & 173.33\footnotesize{$\pm$11.02}                    
\\
                                    & \PVPwoR{}                  & 439.96\footnotesize{$\pm$68.62}        & 1.09\footnotesize{$\pm$2.05}        & 9242.42\footnotesize{$\pm$2056.15}       & 509.45\footnotesize{$\pm$237.64}        & 402.53\footnotesize{$\pm$78.44}      & 6454.75\footnotesize{$\pm$1193.92}       & 891.84\footnotesize{$\pm$251.87}       
                                    % & 8.05\footnotesize{$\pm$6.82}        & -15.76\footnotesize{$\pm$1.51}      & 172.12\footnotesize{$\pm$13.15}
                                    \\
                                    & \PVPwR{}                   & 733.73\footnotesize{$\pm$116.70}       & 14.33\footnotesize{$\pm$3.98}        & 16510.30\footnotesize{$\pm$1684.40}       & 1883.64\footnotesize{$\pm$270.40}        & 1136.07\footnotesize{$\pm$301.34}     & 8777.76\footnotesize{$\pm$1410.76}       & 1097.92\footnotesize{$\pm$246.94}      
                                    % & 28.46\footnotesize{$\pm$0.70}        & 20.11\footnotesize{$\pm$0.97}      & 386.51\footnotesize{$\pm$17.78}
                                    \\
                                    & \BC{}                                     & 537.20\footnotesize{$\pm$51.86}        & 9.09\footnotesize{$\pm$4.64}        & 17273.33\footnotesize{$\pm$2605.40}       & 1348.80\footnotesize{$\pm$236.28}        & \textbf{1711.00}\footnotesize{$\pm$659.06}     & 10252.73\footnotesize{$\pm$1556.18}       & \cellcolor[HTML]{FFFFC7}\textbf{2278.00}\footnotesize{$\pm$337.84}     
                                    % & 28.76\footnotesize{$\pm$0.82}        & -17.94\footnotesize{$\pm$1.19}      & 65.65\footnotesize{$\pm$8.39} 
                                    \\
                                    & \AblaNoTGT{}     & \cellcolor[HTML]{FFFFC7}\textbf{1065.65}\footnotesize{$\pm$164.01}       & 40.56\footnotesize{$\pm$10.22}       & 16465.45\footnotesize{$\pm$1434.87}       & 2785.41\footnotesize{$\pm$452.83}        & 1269.06\footnotesize{$\pm$288.76}     & \textbf{15420.19}\footnotesize{$\pm$2108.08}       & 1676.97\footnotesize{$\pm$652.21}       
                                    % & 32.50\footnotesize{$\pm$0.44}        & 20.72\footnotesize{$\pm$0.19}      & 1146.47\footnotesize{$\pm$45.85}  
                                    \\
                                    & \AblaNoFT{}             & \cellcolor[HTML]{FFFFC7}\textbf{1081.16}\footnotesize{$\pm$131.98}       & 19.35\footnotesize{$\pm$16.53}       &        \cellcolor[HTML]{FFFFC7}\textbf{\color{red}21253.33}\footnotesize{$\pm$2694.61} & \textbf{3037.84}\footnotesize{$\pm$1152.24}       & 891.97\footnotesize{$\pm$148.50}     & 9759.02\footnotesize{$\pm$3093.64}       & 1830.23\footnotesize{$\pm$559.33}       
                                    % & 31.11\footnotesize{$\pm$0.60}        & 20.59\footnotesize{$\pm$0.42}      & 1071.34\footnotesize{$\pm$50.85} 
                                    \\
                                    & \DQfD{}                   & \cellcolor[HTML]{FFFFC7}1003.21\footnotesize{$\pm$128.45}       & 39.27\footnotesize{$\pm$20.85}       & 17003.64\footnotesize{$\pm$1532.05}       & 3146.25\footnotesize{$\pm$418.00}        & 1075.79\footnotesize{$\pm$249.17}     & \textbf{16525.18} \footnotesize{$\pm$1915.10}       & 1952.64\footnotesize{$\pm$427.78}      
                                    % & 32.23\footnotesize{$\pm$0.34        & 20.47\footnotesize{$\pm$0.80      & 1103.69\footnotesize{$\pm$31.51 
                                    \\
                                    & \ours{}                   & \cellcolor[HTML]{FFFFC7}\textbf{\color{red}1167.19}\footnotesize{$\pm$103.02}       & \cellcolor[HTML]{FFFFC7}\textbf{\color{red}61.40}\footnotesize{$\pm$23.91}       & \cellcolor[HTML]{FFFFC7}\textbf{19969.09}\footnotesize{$\pm$1555.69}       & \textbf{\color{red}3368.31}\footnotesize{$\pm$701.54}        & \textbf{\color{red}1736.52}\footnotesize{$\pm$513.91}     & \textbf{\color{red}17862.58}\footnotesize{$\pm$2181.02}       & \cellcolor[HTML]{FFFFC7}\textbf{\color{red}2558.57}\footnotesize{$\pm$331.18}       
                                    % & 31.91 \footnotesize{$\pm$0.66}        & 20.64\footnotesize{$\pm$0.33}      & 1089.27\footnotesize{$\pm$42.73} 
                                    \\
                                    % \cline{2-9}
                                    % & $|\LabelDataE|$ & & & & & & & \\
\bottomrule
\end{tabular}
}
\end{table}

\textbf{\pw{Learning from a combination of two imperfect signals} performs better than learning from $\proxyr$ or corrective actions alone.}
Compared to \Rainbow{}, \Cref{fig:HighwayC1vsGTLess} and \Cref{tab:AtariBaselines} \pw{show} that \ours{} with corrective actions can converge quickly and stably to \pw{the} labeler's performance using proxy rewards with different levels of imprefection, but \Rainbow{}'s performance is unstable and \pw{can easily} converge to non-optimal performance without the guidance from corrective actions.
% Note that even though the \ExpertCL{} is obtained from \Rainbow{} with PRExp, we can not make sure that using the same reward will obtain the desired policy due to the stochastic training.
Compared to \BC{} and \DAgger{}, \Cref{fig:HighwayC1BaselinesLess} and \Cref{tab:AtariBaselines} demonstrate that \ours{}'s performance exceed\pw{s} their performance\pw{s} substantially in most of the games, 
    % or is as good as them in several Atari games, 
    which confirms the effectiveness of integrating the proxy rewards \pw{with the} RL losses to achieve better alignment than using corrective actions alone.

\textbf{\ours{} achieves better alignment performance than baselines.}
As shown in \Cref{fig:HighwayC1BaselinesLess} and \Cref{tab:AtariBaselines}, using the same sample budget in terms of both environmental transitions and label budget, \ours{} achieves the best performance in terms of aligning with labelers' performance in almost all settings, or performs similarly with the best one otherwise.
In \Cref{fig:HighwayC1BaselinesLess}, except for \Rainbow{}, all methods perform similarly in terms of aligning the labeler's \SpeedAvg{} and \LaneChangeAvg{} performance, while for the harder performance metrics \CrashRate{} and \DistanceAvg{}, \ours{} and \DQfD{} perform similarly, and both are significantly better than \PVPwoR{}, \PVPwR{}, and \DAgger{}.
% which 
% \CMT{which?}
In \Cref{tab:AtariBaselines}, with large $\LabelDataE$, \ours{} do\pw{es} match the labelers' performances, except \pw{in} the hardest \hero{} with full action dimensions and relatively long episode length.
Specifically, compared with the two PVP methods, 
    their performance\pw{s are} obviously worse than that of the methods using margin loss, 
    since $\LossPVP$ sets a pre-defined bound for Q-values making it fails to adapt to various reward settings.
In some games, incorporating proxy rewards into their framework (\PVPwR{}) performs even worse than not (\PVPwoR{}).
However, margin loss is a \pw{suitable} choice to incorporate the corrective actions with proxy rewards.
As for \DQfD{}, in simple environments we perform similarly, but in harder Atari environments \ours{} outperforms it and shows more notable performance gaps than in highway.

\textbf{Ablations: \ours{}'s integral design leads to a stable and robust performance.}
Considering that the two-phase learning and pseudo-labeling are the main novel designs in \ours{},
% removing the \PhaseOne-phase and pseudo-labels from \ours{} leads to \DQfD{},
% removing the \PhaseTwo-phase from \ours{} leads to \DAgger{},
there are two additional settings to be examined besides \DQfD{} and \DAgger{}: (1) \AblaNoFT{}, which removes the $\PhaseOne$-phase but retains the pseudo-labels, and (2) \AblaNoTGT{}, which removes pseudo-labels.
In the highway environment, the two ablations perform similarly to \ours{}, and we include them into \Cref{appendix:HighwayMoreResults} instead of \Cref{fig:HighwayC1BaselinesLess} to make it clearer to check.
% \CMT{zj: remember to put them into appendix.}
However, in \Cref{tab:AtariBaselines}, we find that while the two ablation settings may perform well in individual environments when action feedback is abundant, they are not as robust as \ours{} across different environments, especially when feedback is limited and tasks are challenging. \ours{} consistently demonstrates superior performance in such scenarios, maintaining its lead even with fewer feedback instances.
% However, in \Cref{tab:AtariBaselines}, we find that their performances may not be bad in some games but can not be as robust as \ours{} across different environments.
% Combining both design choices together, \ours{} achieves stable alignment across various environments.

\subsection{Overcoming Non-optimality of Corrective Actions with Proxy Rewards}\label{subsec:rand_actions}

\begin{figure}[t]
    \begin{subfigure}{\textwidth}
      \centering
      \includegraphics[width=\linewidth]{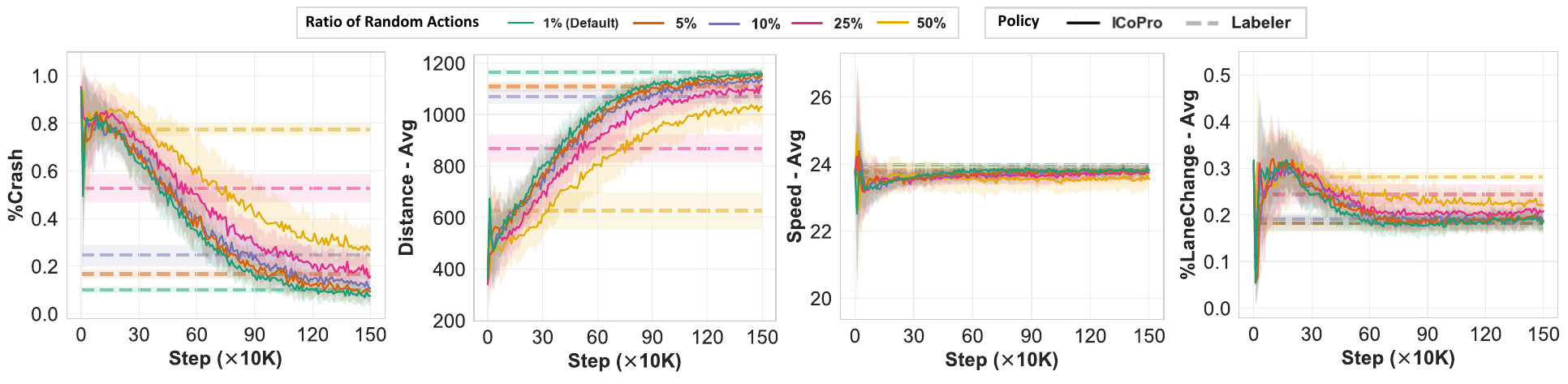}  
      \caption{Highway with \DiffRand{}. Performances are averaged over the set of proxy rewards in \Cref{tab:HighwayReward1ChangeLane}.}
      \label{fig:HighwayC1DiffRandAvgRow}
    \end{subfigure}
    \newline
    \begin{subfigure}{0.5\textwidth}
      \centering
      \includegraphics[width=\linewidth]{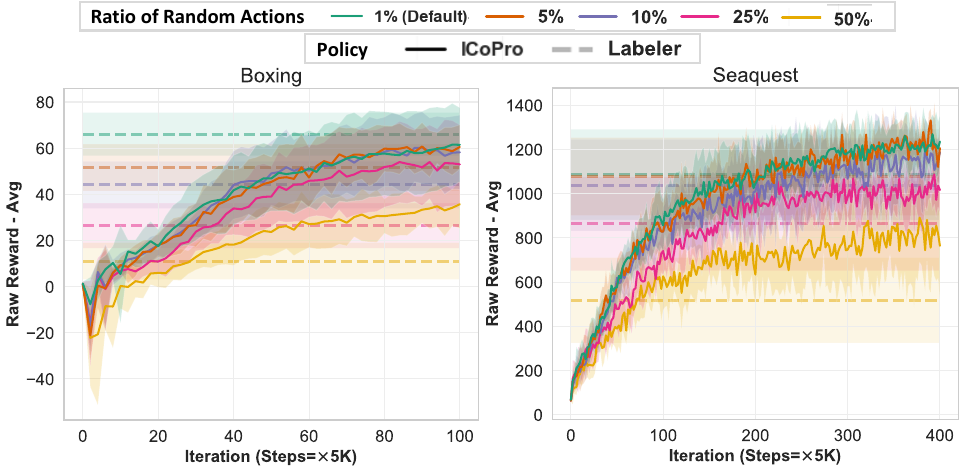}  
      \caption{Atari (Boxing, Seaquest) with \DiffRand{}.}
      \label{fig:Atari_diff_rand_avg}
    \end{subfigure}
    \begin{subfigure}{0.5\textwidth}
      \centering
      \includegraphics[width=\linewidth]{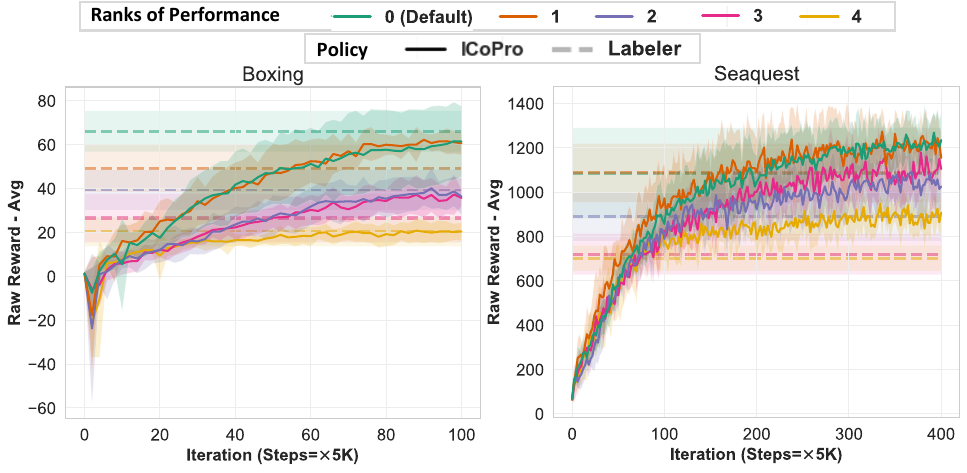}  
      \caption{Atari (Boxing, Seaquest) with \DiffCkpt{}.}
      \label{fig:Atari_diff_ckpt_avg}
    \end{subfigure}
    % \begin{subfigure}{0.5\textwidth}
    %   \centering
    %   \includegraphics[width=\linewidth]{images/}  
    %   \caption{Atari with \DiffCkpt{}}
    %   \label{fig:HighwayC1BaselinesLess}
    % \end{subfigure}
\caption{\ours{} facing different types of non-optimality of corrective actions (different colors). 
% Dashed horizontal lines indicate labelers' performance.
Performances are averaged over large/small $\LabelDataE$ in \Cref{fig:Atari_diff_rand_avg,fig:Atari_diff_ckpt_avg}.}
\label{fig:non_optimal_labeler}
\end{figure}

To simulate different types of imperfections inside corrective actions, the most natural and commonly adopted \citep{lee2021BPref,luo2024RLIF} one is to replace part of a labeler's corrective actions with random actions (\DiffRand{}).
In our setting, it simulates labelers who know which state-action pairs to correct but provide noisy corrective actions.
We evaluate it in three representative environments that cover various environmental complexities: highway, Boxing and Seaquest.
As shown in \Cref{fig:HighwayC1DiffRandAvgRow,fig:Atari_diff_rand_avg}. \ours{} performs robustly within a reasonable ratio of random actions (i.e., $<25\%$), and can perform significantly better than the labelers with large randomness (i.e., $50\%$).
For Atari, we further consider a harder type that uses different labelers with worse performances (\DiffCkpt{}) to simulate labelers who are suboptimal on both where and how to provide corrective actions.
As shown in \Cref{fig:Atari_diff_ckpt_avg}, \ours{} still demonstrate a strong capability to overcome the non-optimality inside such unreasonable action feedback in most of the cases.
% \CMT{zj: explain diff-ckpt and diff-rand}
Performance of Highway using another labeler, and more plots for Atari can also be checked in \Cref{appendix:MoreExperiments}.

\subsection{User study}\label{subsec:userStudy}
\begin{wrapfigure}{r}{6.1cm} % wraptable needs to be put in the front of text
  \centering
  \includegraphics[width=\linewidth]{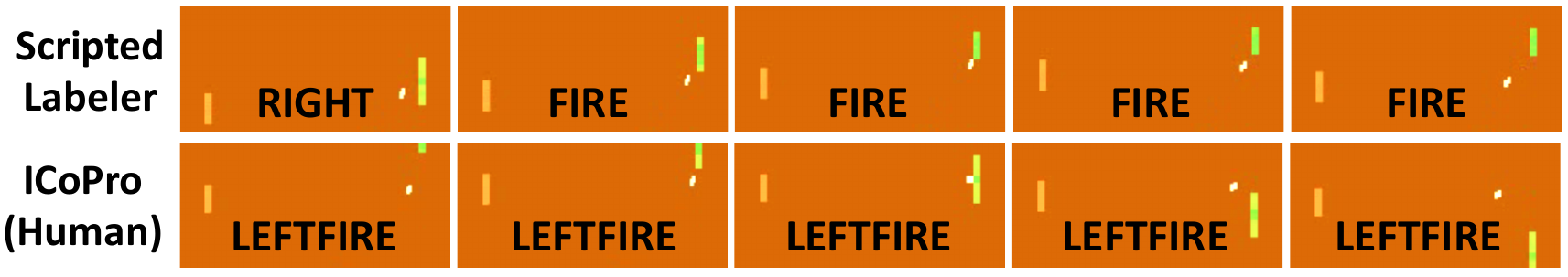}
  % \vspace{-3cm}
  \caption{\pw{Performances} of the \textit{scripted labeler} 
  and \textit{\oursHuman{}} in \pong{}. Each row shows a sequence of state-action pairs. While the scripted labeler prefers to catch the ball with the corner of the paddle, \oursHuman{} prefers to catch the ball with a larger part of the paddle, which is more human-like.}
    \label{fig:PongCompareUserSimu}
\end{wrapfigure}
To verify that \ours{} can align with real human's preference\pw{s when} the desired performance is hard to be obtained with engineered rewards, 
    % ,and (2) have notable better performance than \BC{}.
we use \pong{} and highway as two representative environments to evaluate \textit{\ours{} with real human} (\oursHuman{}) with 500 action labels. 
For \pong{}, \oursHuman{} obtain\pw{s} an agent that performs more human-like than the scripted optimal \zj{labeler}s (see \Cref{fig:PongCompareUserSimu} \pw{for details}).
% For highway, we use the most imperfect one (PR4) as the proxy reward since it does not introduce extra performance shaping, and obtain a vehicle that knows how to take over cars with super fast average speed ([\SpeedAvg{},\CrashRate]=[29.02,0.44]), which is one that we could never obtain with \Rainbow{} using PR4.
For highway, we use PR4 as the proxy reward \pw{in\oursHuman{}} since it does not introduce \pw{any} extra performance shaping, and obtain a vehicle that knows how to take over cars with super fast average speed 
% \begin{wrapfigure}{r}{7cm} % wraptable needs to be put in the front of text
%   \centering
%   \includegraphics[width=0.5\textwidth]{images/user_study/PongCompareUserSimu.pdf}
%   % \vspace{-3cm}
%   \caption{Compare the performance of the \textit{scripted labeler} and \textit{\oursHuman{}} in \pong{}. Each row shows a sequence of state-action pairs. While the scripted labeler prefers to catch the ball with the corner of the paddle, \oursHuman{} prefers to catch the ball with a larger part of the paddle, which is more human-like.}
%     \label{fig:PongCompareUserSimu}
% \end{wrapfigure}
([\SpeedAvg{},\CrashRate]=[29.02, 0.22]), which is \pw{a behavior} that we could never obtain with \Rainbow{} using PR4 \pw{alone}.
Moreover, compared to \oursHuman{}, \BC{} on those real human labels gets poor performance due to the limited data and non-optimality \pw{of human feedback} (Pong: 7.83 vs. -11.26, highway: [29.02, 0.22] vs. [27.90, 0.41])), which confirms again \ours{}'s ability to learn from two imperfect signals in a sample-efficient way.
% \CMT{zj: use a table: \oursHuman{} achieve an average score of 7.83{\footnotesize $\pm$7.43} with only 500K timesteps and 500 action labels, but \BC{}'s is only -11.26{\footnotesize $\pm$4.31}.
% \BC{} on such buffer reach lower \SpeedAvg{} (27.90) and a larger \CrashRate{} (0.82).
 % verifying again that \ours{} is samplecan relieve the non-optimality issu}
More details about our user study and the whole evaluation videos are provided in \Cref{appendix:UserStudy}.
% \CMT{Do we need to multiply by 100 to get the percentage?
% If yes, that's a huge crash percentage, no?
% zj: a mistake, should be divided by 2}

% \CMT{zj: keep this? 
% with the same hyper-parameters as previous experiments but only 100 iterations.
% Human labelers need to provide 500 and 1000 feedback in total on \pong{} and highway, respectively. We let two humans work on the two tasks individually, and each person ran 2 seeds for his/her task.}

\section{Conclusions and limitations}\label{appendix:ConclusionLimitation}
% \CMT{zj: pay attention to the title format}
We present a human-in-the-loop framework where an RL agent can learn from two potentially unreliable signals: corrective actions provided by \pw{a} \zj{labeler} and a pre-defined proxy reward. 
The motivation for combining the two learning signals is threefold: (1) guide the agents' training process with corrective actions, 
(2) use proxy rewards to help reduce human labeling efforts, and
(3) possibly compensate for the imperfection of one signal with the other.
Our value-based method trains a policy in an iterative way, with two separate learning phases inside each iteration:
    the 
    $\phaseOne$-phase uses \pw{a} margin loss to update the policy to better align with action feedback,
    then the $\phaseTwo$-phase incorporates RL losses as well as \pw{the margin loss expressed with} pseudo-labels to generalize the improved values.
    \CMT{Is "generalize" the correct word here?}
Our experiments validate our algorithmic design.
    
Although our method achieves better and robust performance compared with baseline methods, it could be further improved by replacing uniform \zj{query selection} 
% sampling ---a naive way to perform query selection---
by more advanced sampling methods, which could help reduce the number of needed feedback \zj{further}.
In addition
, while we demonstrated the feasibility and the benefit of learning from two imperfect signals, we \pw{plan} as future work to provide a more theoretical analysis
% , which could 
\zj{to}
reveal what assumptions about the misspecification of proxy rewards and/or the suboptimality of the corrective actions could guarantee their 
% \pw{effective}
synergetic 
combination.
% \CMT{We need an adjective to combination, because we can also combine them, but we want to ensure that their combination provides a better performance. zj: previously I delete this word to make the paragraph to be one line shorter. So we could keep using ``synergetic'' instead of ``effective''?}

% \Note{Page limit is 10.
% \href{https://iclr.cc/Conferences/2025/CallForPapers}{https://iclr.cc/Conferences/2025/CallForPapers} said: ``New this year, the main text must be between 6 and 10 pages (inclusive). This limit will be strictly enforced. Papers with main text on the 11th page will be desk rejected. The page limit applies to both the initial and final camera ready version.
% We encourage authors to be crisp in their writing by submitting papers with 9 pages of main text. We recommend that authors only use the longer page limit in order to include larger and more detailed figures. However, authors are free to use the pages as they wish, as long as they obey the page limits.''
% }
\bibliography{iclr2025_conference}
\bibliographystyle{iclr2025_conference}

\newpage
\appendix

\section{More experimental results and plots}\label{appendix:MoreExperiments}
\subsection{Atari}\label{appendix:MoreExperimentsAtari}
\subsubsection{Learning Curves for \Cref{tab:AtariBaselines} Including Pong, Freeway, and Enduro}
\begin{figure}[ht]
    \centering
    \includegraphics[width=\textwidth]{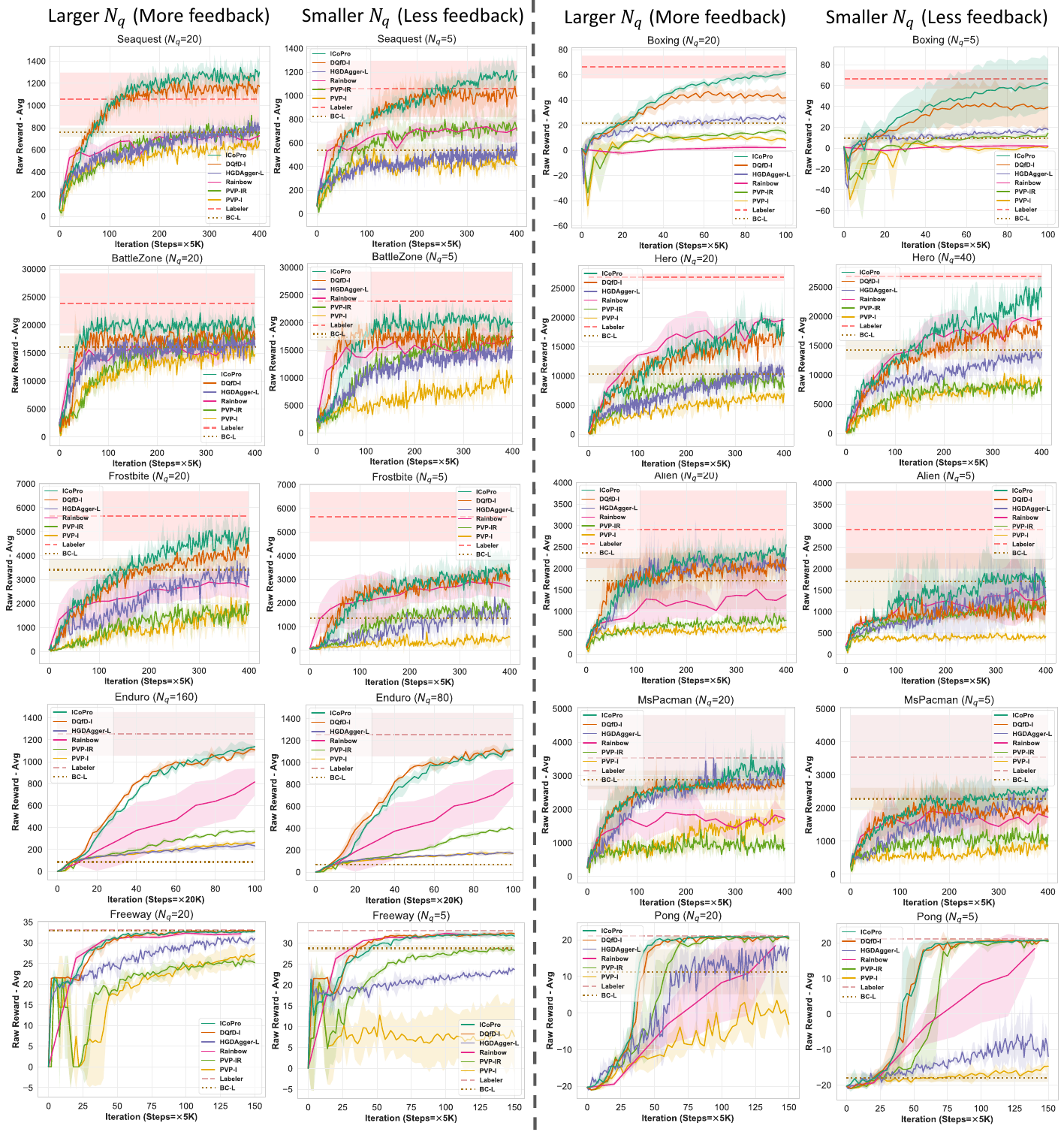}
    \caption{Compare baseline methods on Atari in terms of the averaged episode return measured with the raw reward $\rawrAtari$. The shadow indicates the standard deviation over 5 seeds. $\CntQueryItr$ in titles refer to the number of queries per iteration, and the larger (resp. smaller) ones correspond to the large (resp. small) $\LabelDataE$ in \Cref{tab:AtariBaselines}.}
    \label{fig:AtariBaselines}
\end{figure}

\begin{figure}[ht]
    \centering
    \includegraphics[width=\textwidth]{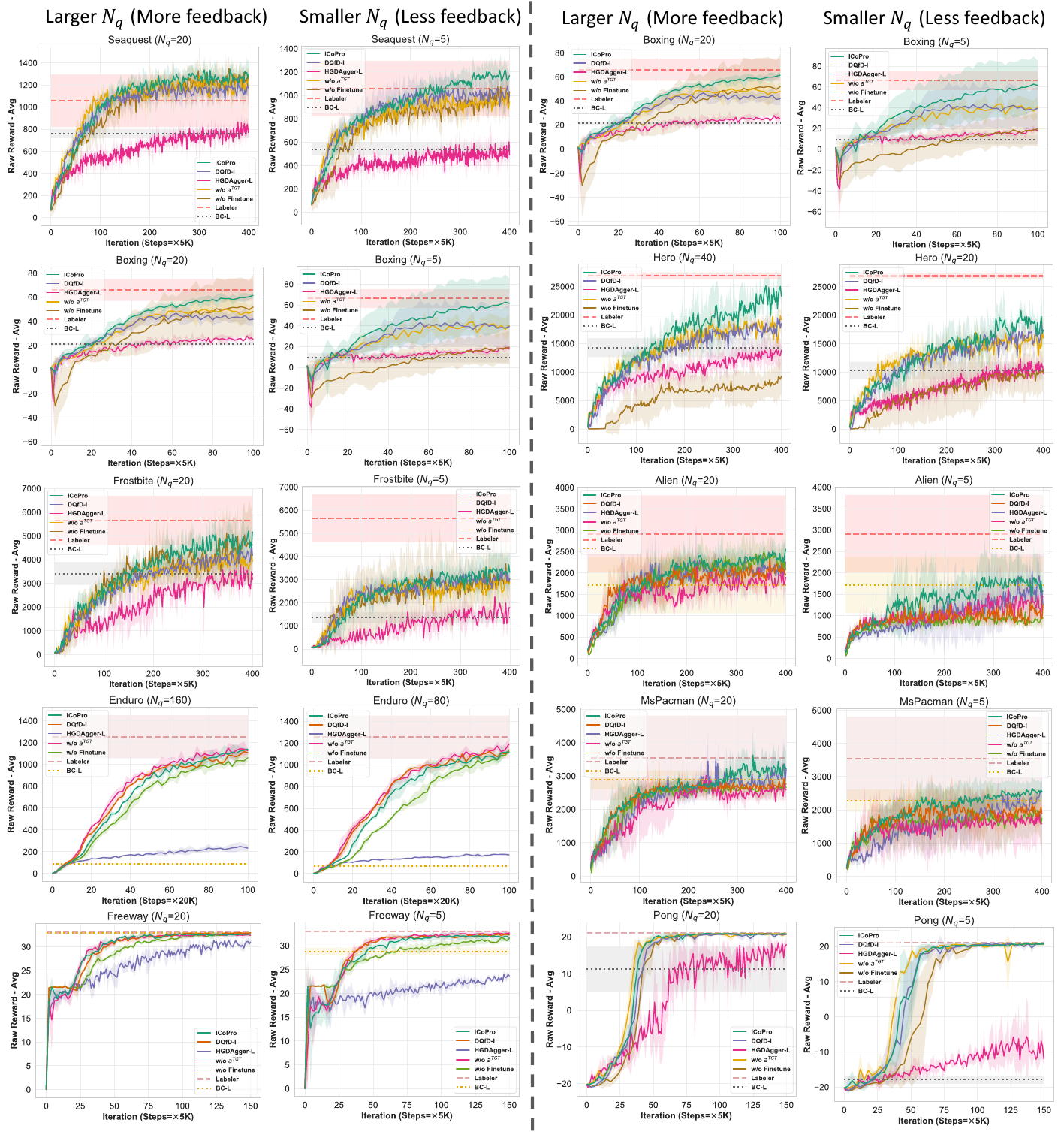}
    \caption{Ablation study on Atari to test the effect of \ours{}'s two-phase scheme and target pseudo label. $\CntQueryItr$ in titles refer to the number of queries per iteration, and the larger (resp. smaller) ones correspond to the large (resp. small) $\LabelDataE$ in \Cref{tab:AtariBaselines}.}
    \label{fig:AtariAblaBasic}
\end{figure}

\Cref{fig:AtariBaselines} show plots comparing \ours{} with various baselines, 
 and \Cref{fig:AtariAblaBasic} show ablation study for \ours{}. 
Lines represent the average episodic return in terms of the raw reward. Shadows represent the standard deviation.
For Rainbow, the x-axis should read in terms of steps instead of iterations.
For other methods, the x-axis can be read in terms of steps or iterations.

% \TODO{explaination for battlezone: feedback is strange, with smaller amount of feedback, the performance is better, and BC best score is achieved when acc=0.6 or 0.7; and this is also observed in ablation that }

% When we evaluate \BC{}'s performance on \battlezone{}, we find that the best performance is obtained with a wired low prediction accuracy (about 0.6), which can explain why the learning curve for this game in \Cref{fig:AtariBaselines} and \Cref{fig:AtariAblaBasic} converge to the lower performance of the \zj{labeler}'s.

\subsubsection{Compare \ours{} with More Related Works.}\label{appendix:CompareOtherMethods}
\paragraph{Compare with \citet{ilhan2021ActReuseImitation}.}
In \Cref{tab:compareActionReuse}, we compare the performance of \ours{} with ActionImitation \citep{ilhan2021ActReuseImitation}, which is a method that also uses action advice but is to augment the replay buffer, on their experimental environments.
Results for ActionImitation come from their paper.
Our simulated \zj{labeler}s in the three games achieved the same score as theirs (full score on \pong{} and \freeway{}, and 1200 score on \ed{}), so we put \ours{}'s results from \Cref{fig:AtariBaselines} directly to compare with.
Our method uses significantly less data than the ActionImitation method.

\begin{table}[ht]
    \centering
    \caption{Compare the number of feedback actions ($|\LenLabelData|$) and environmental interaction timesteps ($T_{env}$) needed to reach the \zj{labeler}s' score in each environment.}
    \label{tab:compareActionReuse}
    \begin{tabular}{l|ccc|ccc}
    \toprule
     \multirow{2}{*}{Method}& \multicolumn{3}{c|}{$|\LenLabelData|$}& \multicolumn{3}{c}{$T_{env}$} \\
      & \pong{} & \freeway{} & \ed{} & \pong{} & \freeway{} & \ed{}\\
     \midrule
      ActionImitation & 10K & 10K & 10K & 3M & 3M & 3M\\
      \ours{} & 375 & 2K & 8k & 375K & 500K & 2M\\
      \bottomrule
    \end{tabular}
\end{table}
% \paragraph{Compare \ours{} with \citet{luo2024RLIF} using \DiffRand{}.}
% \citet{luo2024RLIF}
\subsubsection{Detailed Results for \DiffRand{} and \DiffCkpt{} with Different Label Budget Size}
\begin{figure}[t]
    \begin{subfigure}{0.49\textwidth}
      \centering
      \includegraphics[width=\linewidth]{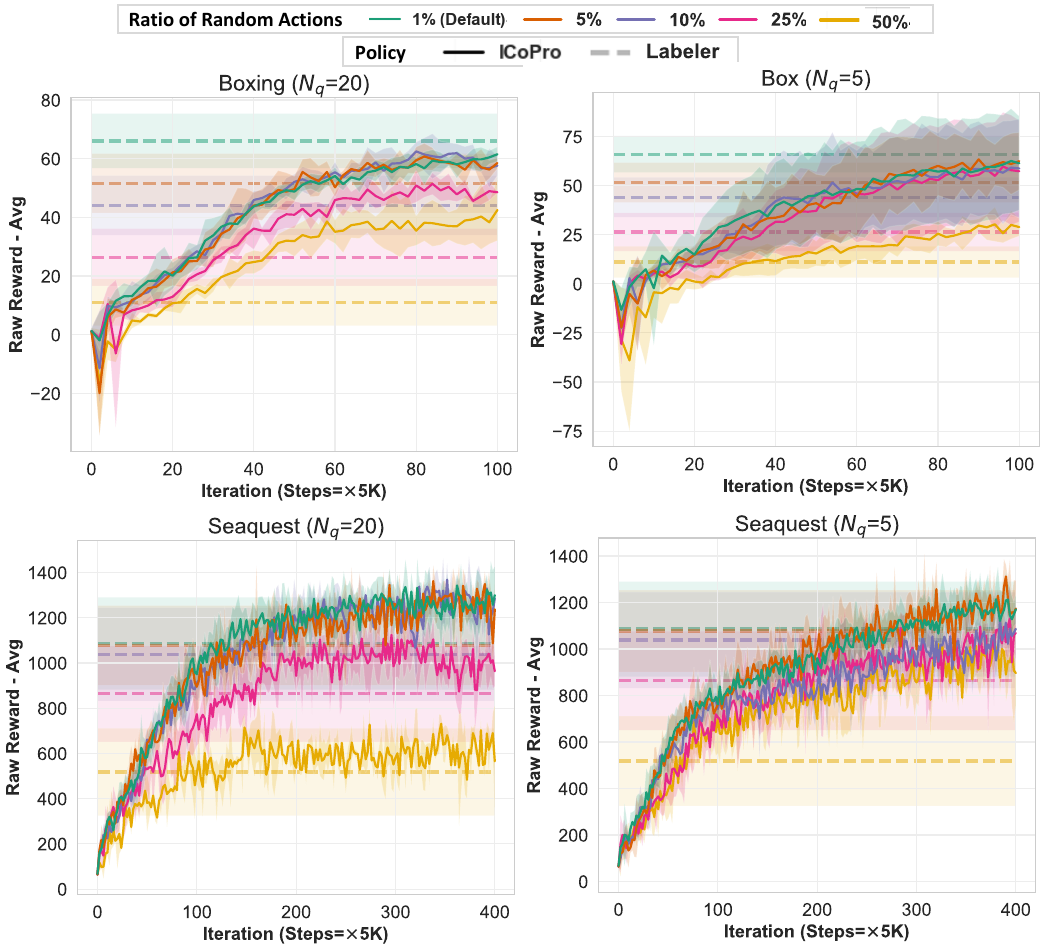}  
      \caption{\DiffRand{} correspond to \Cref{fig:Atari_diff_rand_avg}.}
      \label{fig:Atari_diffrand}
    \end{subfigure}
    % \newline
    \begin{subfigure}{0.49\textwidth}
      \centering
      \includegraphics[width=\linewidth]{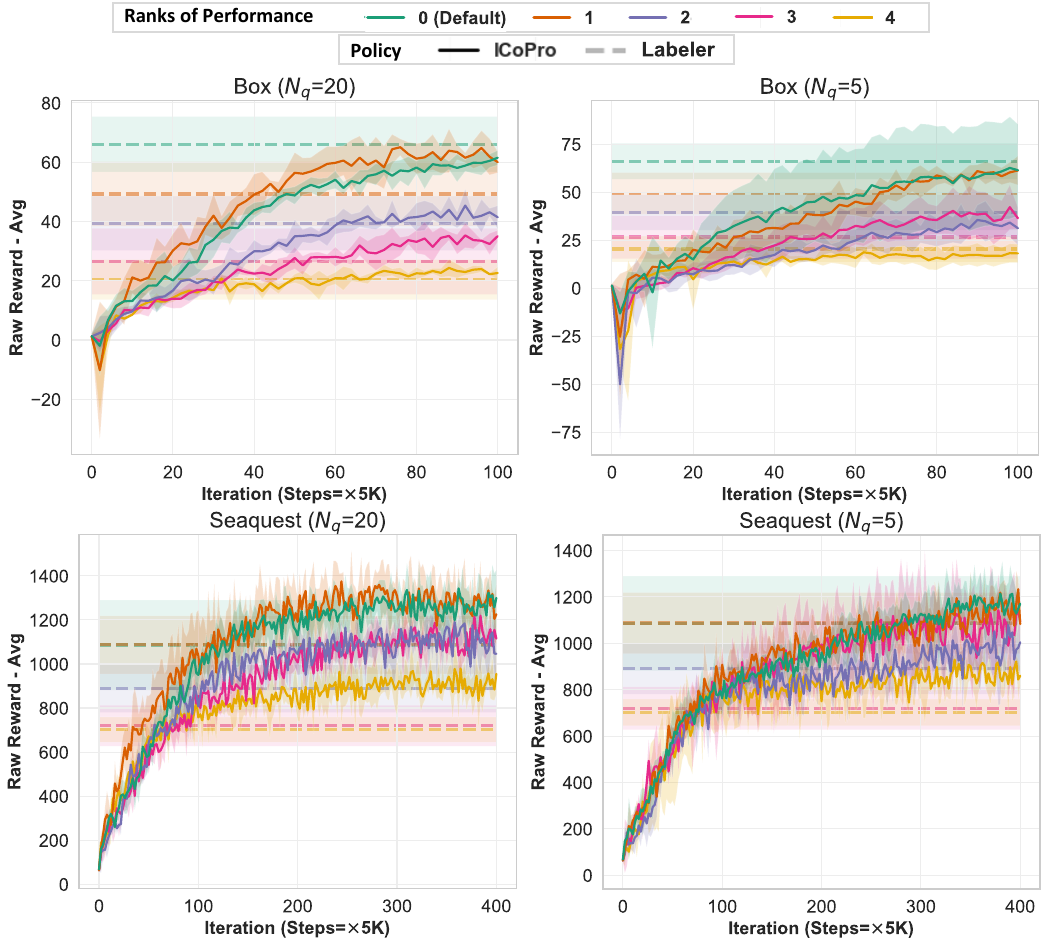}  
      \caption{\DiffCkpt{} correspond to \Cref{fig:Atari_diff_ckpt_avg}.}
      \label{fig:Atari_diffckpt}
    \end{subfigure}
\caption{Detailed plots with large/small budget size for the averaged plots in \Cref{fig:Atari_diff_rand_avg}. Larger (resp. smaller) $\CntQueryItr$ indicates the large (resp. small) budget $\LabelDataE$.}
\label{fig:Atari_non_optimal_labeler}
\end{figure}

\Cref{fig:Atari_non_optimal_labeler} are separated plots using different size of label busget in \Cref{fig:Atari_diff_rand_avg,fig:Atari_diff_ckpt_avg}.
For \DiffCkpt{}, a larger number of imperfect labels tends to lead to a worse performance, as shown in \Cref{fig:Atari_diffckpt}.
But for \DiffRand{}, a larger number of imperfect labels tends to lead to performance with less variance but may not necessarily to a worse performance, as shown in \Cref{fig:Atari_diffrand}.

\subsection{Highway}\label{appendix:HighwayMoreResults}
\subsubsection{Extra Results for Ablation Study}\label{appendix:HighwayMoreAblations}
\begin{figure}[ht]
  \centering
  \includegraphics[width=\textwidth]{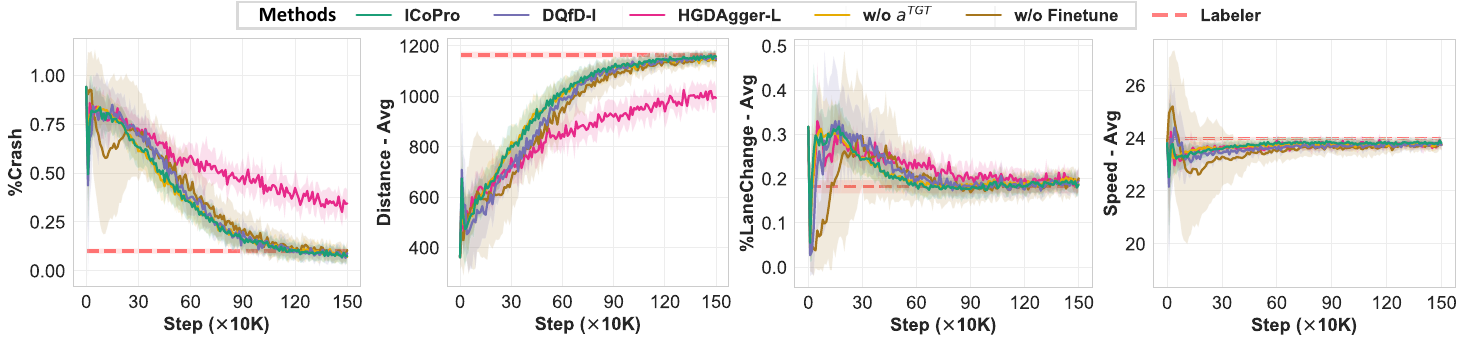}  
  \caption{Extra ablation results on highway with the labeler used in \Cref{fig:highwayCkpt1ChangeLane}. Performances are averaged over the set of proxy rewards mentioned in \Cref{tab:HighwayReward1ChangeLane}.}
  \label{fig:HighwayC1Ablations}
% Feedback schedule is $\CntQueryItr=10$, $\LenRollout=1k$, $\TotalIter=150$, 
\end{figure}

\Cref{fig:HighwayC1Ablations} shows the ablation study on highway. For such an environment with a relatively small number of action dimensions, \ours{} can outperform \DAgger{} significantly but only outperform other ablations slightly. Similar performance can also be observed in Atari games with small action dimension like Pong and Freeway in \Cref{fig:AtariAblaBasic}.

\subsubsection{Extra Evaluation with Another Scripted Labeler (\DiffCkpt{})}\label{appendix:HighwayAnotherCkpt}
In \Cref{tab:ExpertPerformance}, we compare the two simulated \zj{labeler}s used in our experiments.
The first \zj{labeler} (\ExpertCL{})\zj{, which is the labeler illustrated in our main text,} prefers driving faster and changing to different lanes more often than the second one, but with a larger crash rate as well.
The second one (\ExpertRL{}) drives in a safer way at a lower speed and prefers to drive in lanes on the right.
Then, in \Cref{tab:HighwayReward2RightLane}, we list the 2 sets of proxy rewards for the two \zj{labeler}s.

\begin{figure}[ht]
    \begin{subfigure}{\textwidth}
      \centering
      \includegraphics[width=\linewidth]{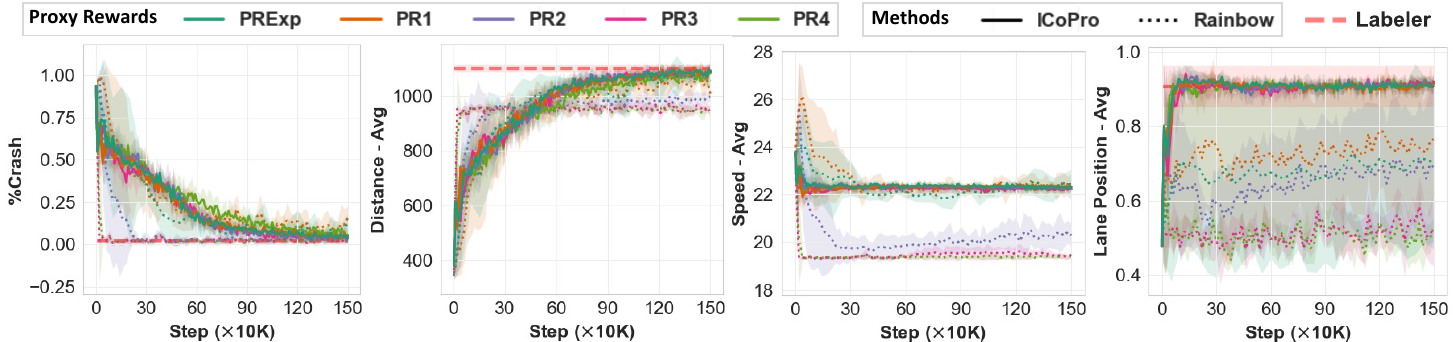}  
      \caption{Compare agent's performance learned from \ours{} and Rainbow.}
      \label{fig:HighwayC2vsGTLess}
    \end{subfigure}
    \newline
    \begin{subfigure}{\textwidth}
      \centering
      \includegraphics[width=\linewidth]{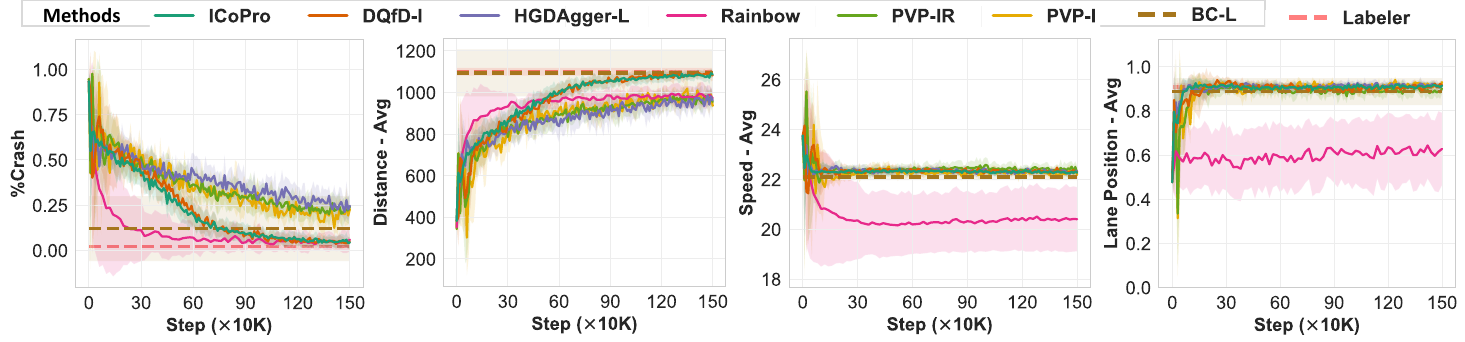}  
      \caption{Compare \ours with other baselines. The learning curves are averaged over the set of proxy rewards mentioned in \Cref{tab:HighwayReward2RightLane}.}
      \label{fig:HighwayC2BaselinesLess}
    \end{subfigure}
\caption{Experiments on highway with \ExpertRL{} over the set of proxy rewards mentioned in \Cref{tab:HighwayReward2RightLane}. Each plot compares the performance with respect to one representative performance metric. Compared with \Cref{fig:highwayCkpt1ChangeLane}, the different performance metric is \LanePosAvg{} in the 4th subplot.
% Feedback schedule is $\CntQueryItr=10$, $\LenRollout=1k$, $\TotalIter=150$, 
$|\LabelDataE|$=1.5K at the end.}
\label{fig:highwayCkpt2RightLane}
\end{figure}

\Cref{fig:highwayCkpt2RightLane} shows experimental results with \ExpertRL{}, which confirm our analysis in \Cref{subsec:exp_one_ckpt}
 again that \ours{} can align to the desired performance better than baselines.
 
\begin{table}[ht]
\centering
\caption{Performance metric for the 2 simulated \zj{labeler}s in highway.}
\label{tab:ExpertPerformance}
\scalebox{0.73}{
    \setlength\tabcolsep{2.5pt} % smaller space between columns
    \begin{tabular}{lcccccccc}
    \toprule
    \zj{Labeler} & \CrashRate{} & \StepAvg{} & \StepMin{} & \DistanceAvg{} & \DistanceMin{} & \SpeedAvg{} & \LanePosAvg{} & \LaneChangeAvg{} \\
    \midrule
    \ExpertCL{}   & 0.10             & 48.51\footnotesize{$\pm$0.39}            & 20.80\footnotesize{$\pm$9.78}            & 1163.689\footnotesize{$\pm$12.85}            & 483.09\footnotesize{$\pm$227.80}            & 23.97\footnotesize{$\pm$0.06}            & 0.57\footnotesize{$\pm$0.05}              & 0.18\footnotesize{$\pm$0.02}   \\
    \ExpertRL{}    & 0.02 & 49.29\footnotesize{$\pm$0.74}           & 28.95\footnotesize{$\pm$19.92}           & 1099.38\footnotesize{$\pm$19.02}            & 608.93\footnotesize{$\pm$404.97}            & 22.30\footnotesize{$\pm$0.12}            & 0.91\footnotesize{$\pm$0.02}             & 0.04\footnotesize{$\pm$0.01}   \\
    \bottomrule
    \end{tabular}
}
\end{table}

\begin{table}[ht]
    \centering
    \caption{Reward weights for engineered proxy rewards on highway for the two \zj{labeler}s mentioned in \Cref{tab:ExpertPerformance}.}
    \label{tab:HighwayReward2RightLane}
    \scalebox{0.85}{
    \begin{tabular}{l|l|ccccc}
        \toprule
        \multicolumn{7}{c}{\ExpertCL{}}\\
        \midrule
        \multicolumn{2}{c|}{Proxy Rewards}  & PRExp & PR1 & PR2 & PR3 & PR4\\
        \midrule
       \multirow{5}{*}{Evnets} & Change lane action & 0.2 & 0 & 0.2 & 0 & 0\\
        & Normalized lane index & 0  & 0 & 0 & 0 & 0\\
        & High speed & 1.5 & 2 & 0.8 & 0 & 0\\
        & Low speed & -0.5 & -1 & 0 & 0 & 0\\
        & Crash & -1.7 & -1 & -1 & -1 & -1\\
        \midrule
        \multicolumn{2}{c|}{Normalize} & [-1,1] & [-1,1] & [-1,1] & [-1,1] & None\\
        \toprule
        \multicolumn{7}{c}{\ExpertRL{}}\\
        \midrule
        \multicolumn{2}{c|}{Proxy Rewards}  & PRExp & PR1 & PR2 & PR3 & PR4\\
       \midrule
        \multirow{5}{*}{Evnets} & Change lane action & 0.2 & 0 & 0 & 0 & 0\\
        & Normalized lane index & 0.5  & 0.5 & 0.2 & 0 & 0\\
        & High speed & 1.7 & 1.5 & 0.8 & 0 & 0\\
        & Low speed & -0.5 & -0.5 & 0 & 0 & 0\\
        & Crash & -1.9 & -1.5 & -1 & -1 & -1\\
         \midrule
        \multicolumn{2}{c|}{Normalize} & [-1,1] & [-1,1] & [-1,1] & [-1,1] & None\\
        \bottomrule
    \end{tabular}
    }
\end{table}

\subsubsection{Extra Results for \DiffRand{}}
\begin{figure}[ht]
      \centering
      \includegraphics[width=\textwidth]{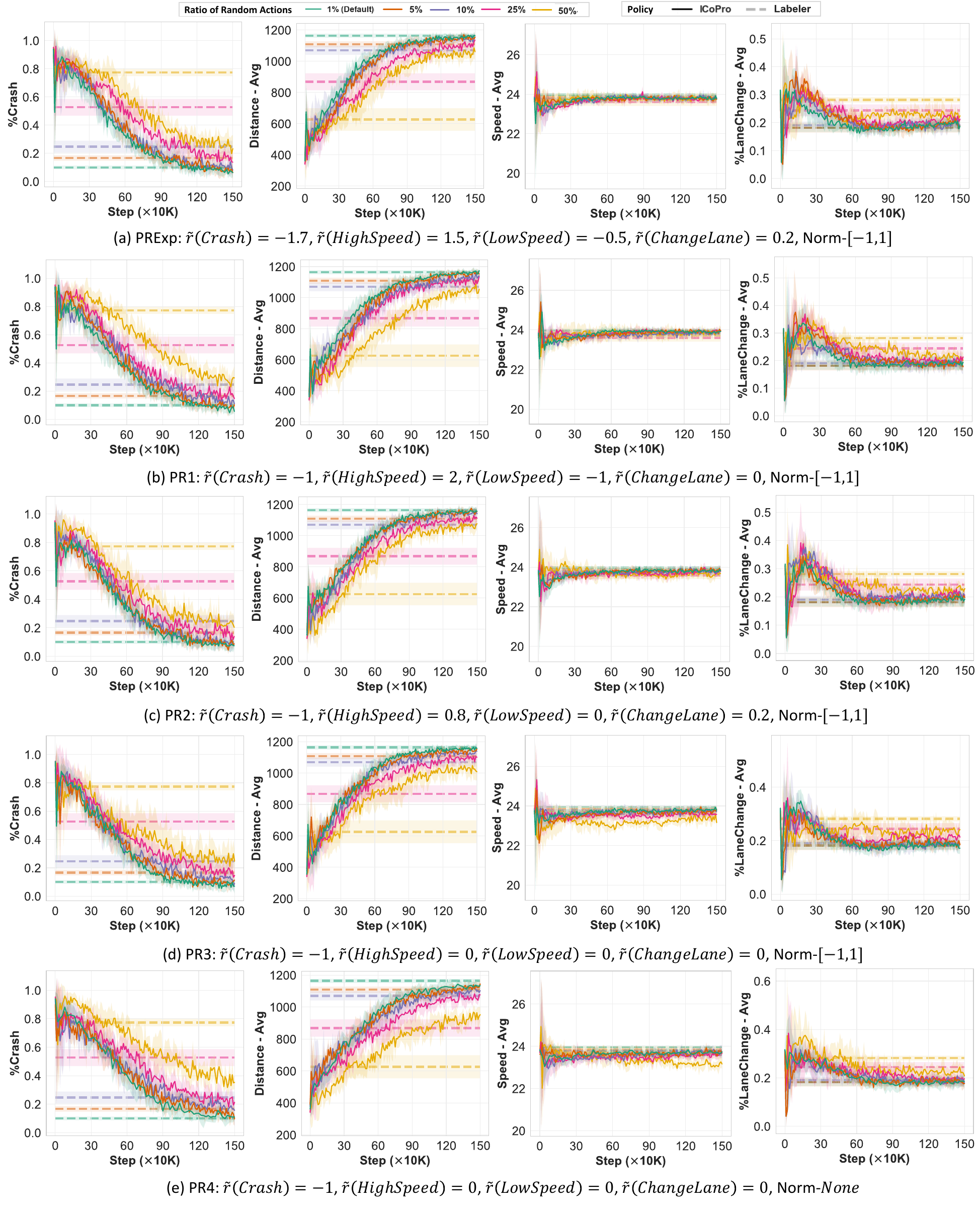}  
      \caption{Detailed performance for each proxy rewards that used to show the average performance in \Cref{fig:HighwayC1DiffRandAvgRow}.}
      \label{fig:highway_diffproxy_diffrand}
\end{figure}

In \Cref{fig:highway_diffproxy_diffrand} we show the whole detailed plots for the averaged ones mentioned in \Cref{fig:HighwayC1DiffRandAvgRow}. \ours{} overcome such non-optimality inside the non-optimal corrective actions no matter which proxy reward is applied.

\section{Implementation details for \ours{}}\label{appendix:Implementatil}
\subsection{Pseudo-code for \ours{}}\label{appendix:PseeudoCodes}
\Cref{algo:ours} show the concrete learning procedure of \ours{}.
\begin{algorithm}[h!]
    \caption{\ours{}}
    \label{algo:ours}
    \begin{algorithmic}[1]
        \Require Initial policy $\pia_1$ with randomly initialized $\Qa_1$,
                 oracle policy $\piE$ with $\QE$
        \Require \# training iteration $\TotalIter$,
                 \# rollout steps $\LenRollout$ per iteration,
                 \# queries per iteration $\CntQueryItr$
        \Require In $\phaseOne$-phase: Target Accuracy $\AccTarget$
        \Require In $\phaseTwo$-phase: \# update epochs $\RLEpo$, data from the recent $\RLIter$ iterations
        \Require $\LabelDataE\leftarrow\emptyset$
        \For{$\CurrentIter \gets 1$ to $\TotalIter$}
            \State Collect a rollout $\EnvData_\CurrentIter=\{(s_t^\CurrentIter,a_t^\CurrentIter,\widetilde{r}_t)\}_{t=1}^{\LenRollout}$ with $a_t^\CurrentIter=\EpsG(\arg\max_{a\in\Ac}\Qa_\CurrentIter(s_t^\CurrentIter,\cdot))$
            \State Sample segments $\mathcal{D}^{\Query}_\CurrentIter=\{\Query_k\}_{k=1}^{\CntQueryItr}$ from $\EnvData_i$
            \For {each $\Query\in\mathcal{D}^{\Query}_\CurrentIter$} 
                \Comment{\PhaseData-phase}
                \State Give corrective actions $\LabelDataE_i=\{(s_t^\CurrentIter, a_{E,t}^\CurrentIter)_k\}_{k=1}^{\CntCFSeg}$ 
                % to the top $\CntCFSeg$ states based on $Q^{diff}(s,a,a^E)$
                % $Q^E(s_t^i,a_{E,t}^i)-Q^E(s_t^i,a_t^i)$
                \State $\LabelDataE\leftarrow\LabelDataE\cup\LabelDataE_i$
            \EndFor
            \State Reset optimizer's 1st and 2nd order moment
            \While {$\Expect_{s\sim\LabelDataE}\left[\mathbb{I}\left[a^E=\argmax_a\Qa_k(s,\cdot)\right]\right]<\AccTarget$}  \Comment{$\PhaseOne$-phase}
                \State Update $\Qa_{\CurrentIter}$ with $\LossMargE$ (\Cref{equ:LossLabel}) on minibatch from $\LabelDataE$
            \EndWhile
            \State Reset optimizer's 1st and 2nd order moment
            \For {$j \gets 1$ to $\RLEpo$} \Comment{$\PhaseTwo$-phase}
                \State $\QTarget\leftarrow \Qa_{\CurrentIter}$ 
                \For{minibatchs $\overline{\EnvData}$ in $\EnvData_{\zj{[\CurrentIter-\RLIter, \CurrentIter]}}$}
                    \State Calculate $\LossOneTD$, $\LossNTD$, and $\LossMargTGT$ on $\overline{\EnvData}$
                    \State Calculate $\LossMargE$ in $\overline{\LabelDataE}\sim\LabelDataE$, $|\overline{\LabelDataE}|=|\overline{\EnvData}|$
                    \State Update $\Qa_{\CurrentIter}$ with $\LossPropETGT$ (\Cref{equ:LossPropETGT})
                \EndFor
            \EndFor
            \State $\Qa_{\CurrentIter+1}\leftarrow\Qa_{\CurrentIter}$
        \EndFor\\  
        \Return {$\Qa_{\TotalIter}$}
    \end{algorithmic}
\end{algorithm}
% epsilon for \zj{labeler} and training agent's evaluation

\subsection{Hyper-parameters}\label{appendix:HyperParameters}
The basic hyper-parameters for \ours{} in all environments are listed in \Cref{tab:BasicHyperparam}.
\begin{table}[ht!]
    \caption{Default hyper-parameters.}
    \label{tab:BasicHyperparam}
    \centering
    \scalebox{0.9}{
    \setlength\tabcolsep{2.5pt}
    \begin{tabular}{lll}
    \toprule
                               & Hyper-parameter & Value \\
    \midrule
    \multirow{3}{*}{General} & training batch size $\bs$ & 128 \\
                             & margin $\MarginConst$ & 0.05 \\
                            & Architecture for Neural Network & same with DERainbow\citep{van2019DERainbow}\\
    \midrule
    \multirow{4}{*}{Optimizer} & type & Adam\citep{diederik2014Adam} \\
                               & learning rate $\lr$ & 0.0001 \\
                               &  eps & $0.01/\bs$ \\
                               & betas & (0.9, 0.999) \\
    \midrule
    \multirow{4}{*}{$\PhaseData$-phase} & $N_{CF}$ & 1 \\
                                      & $\epsilon$ & 0.01 \\
                                      & $\LenSeg$ & Atari: 25, highway: 10\\
                                      & query sampling & uniformly sample segments without overlap\\
    \midrule
    \multirow{1}{*}{$\PhaseOne$-phase} & Accuracy target $\AccTarget$ & 0.98 \\
                               % & maximal number of updates per iteration & 6000 \\
    \midrule
    \multirow{3}{*}{$\PhaseTwo$-phase} & discount factor $\gamma$ & 0.99 \\
                               % & maximal number of updates per iteration & 6000 \\
                               & weight for $\LossMargTGT$: $\WLossMargTGT$ & 0.5 \\
                               & weight for $\LossMargE$: $1-\WLossMargTGT$ & 0.5 \\
     
    \bottomrule
    \end{tabular}
    }
\end{table}

In \Cref{tab:HypersDiffEnv}, we list hyper-parameters with slight differences in different environments. 
The reasons that we do not use a same configuration is:
\begin{itemize}
    \item Choices of $\StepN$ \zj{in \Cref{equ:NStepTDLoss}} are the same one as training the Rainbow \zj{labeler}.
    We find that in \boxing{} and \ed{} we can not obtain meaningful checkpoints with the default value 20 and therefore using smaller values.
    \item $\RLEpo$ defaults to 2, except for the 3 hard Atari games with 18 actions. 
     We observed that setting $\RLEpo=2$ will be hard to let \ours{} and \DQfD{} reach \zj{labeler}s' performance, but setting $\RLEpo=1$ performs better.
     \item Since \ed{}'s episode length can reach 25K when the performance approach to our \zj{labeler}s'.
     In this case, set $\LenRollout$ as the default 5K can introduce bias in our iterative feedback schedule. Therefore we set $\LenRollout=20K$ in this game.
\end{itemize}

% - About the replay ratio, although it seems that we choose a very large replay ratio and som previous works show that a large replay ratio will lead to the loss of plasticity and destroy the whole performance, but note that the replay ratio is in fact gradually increased since at the beginning we do not have enough data from all $Itr$ iterations.

Moreover, we compare the design elements between \ours{} and the original Rainbow's \citep{hessel2018Rainbow} in \Cref{tab:compareDQNs}.

\begin{table}[ht]
    \centering
    \caption{Specific configurations and hyper-parameters for each environment. \zj{Checkpoint}-step is the timestep to obtain our simulated \zj{labeler}s (i.e., policy checkpoints). Meanings for other configurations are consistent with \Cref{algo:ours}. The total environmental steps equals to $\TotalIter\times\LenRollout$, $|\LabelDataE|$ equals to $\TotalIter\times\CntQueryItr\times\CntCFSeg$, $|\EnvData_{[\CurrentIter-\RLIter, \CurrentIter]}|$ equals to $\RLIter\times\LenRollout$.}
    \label{tab:HypersDiffEnv}
    \scalebox{0.66}{
    \setlength\tabcolsep{2.5pt} % smaller space between columns
    \begin{tabular}{l|cccccccccc|cc}
        \toprule
        \multirow{2}{*}{Configuration} & \multicolumn{10}{c}{Atari-Games} & \multicolumn{2}{c}{Highway-\zj{Labeler}s} \\
         & \seaquest{} & \boxing{} & \battlezone{} & \frostbite{} & \alien{} & \hero{} & \mspacman{} & \freeway{} & \pong{} & \ed{} & \ExpertCL{} & \ExpertRL{} \\
        \midrule
        \zj{Checkpoint}-step  & 4.6M    & 10M    & 4.6M       & 4.6M      & 5M    & 5M   & 4M       & 2M      & 1.8M & 4.8M   & 330K              & 990K              \\
        % Eval-step  & 2M    & 500k    & 2M       & 2M      & 2M    & 2M   & 2M       & 750K      & 750K & 2M   & 150K              & 150K              \\
        Step $\StepN$  & 20      & 10     & 20         & 20        & 20    & 20   & 20       & 20      & 20   & 3      & \multicolumn{2}{c}{20}                \\
        $\RLEpo$ & 1       & 2      & 1          & 1         & 2     & 2    & 2        & 2       & 2    & 2      & \multicolumn{2}{c}{2}                 \\
        $\RLIter$ & 64 & 64 & 64 & 64 & 64 & 64 & 64 & 64 & 64 & 16 & \multicolumn{2}{c}{100}\\ 
        $\LenRollout$  & 5K      & 5K     & 5K & 5K        & 5K    & 5K   & 5K       & 5K      & 5K   & 20K    & \multicolumn{2}{c}{1K}     \\
        $|\EnvData_{[\CurrentIter-\RLIter, \CurrentIter]}|$ & 32K & 32K & 32K & 32K & 32K &32K & 32K & 32K & 32K & 32K   & \multicolumn{2}{c}{100K}\\ 
        $\CntCFSeg$ & \multicolumn{10}{c|}{1} & \multicolumn{2}{c}{1}\\ 
        Small $\CntQueryItr$ ($|\LabelDataE|$) & 5 (2K) & 5 (500) & 5 (2K) & 5 (2K) & 5 (2K) & 20 (8K) & 5 (2K) & 5 (750) & 5 (750) & 80 (8k) & \multicolumn{2}{c}{10} \\
        Large $\CntQueryItr$ ($|\LabelDataE|$) & 20 (8K) & 20 (2K) & 20 (8K) & 20 (8K) & 20 (8K) & 40 (16K) & 20 (8K) & 20 (8K) & 20 (3K) & 160 (16K) & \multicolumn{2}{c}{10} \\
        $\TotalIter$ & 400 & 100 & 400 & 400 & 400 & 400 & 400 & 150 & 150 & 100 & \multicolumn{2}{c}{150}\\ 
        % $|\EnvData|$ & 32K & 32K & 32K & 32K & 32K &32K & 32K & 32K & 32K & 32K   & \multicolumn{2}{c}{100K}\\ 
        Total Env Steps & 2M & 500k & 2M & 2M & 2M & 2M & 2M & 750K & 750K & 2M & \multicolumn{2}{c}{150K}\\ 
        \bottomrule
    \end{tabular}
    }
\end{table}

\subsection{Experiments Compute Resources}\label{appendix:Experiments Compute Resources}
For experiments in Atari, we only need one GPU card to launch experiments, including
    GPUs  GeForce RTX 3060 12G GPU + 48GB memory + Intel Core i7-10700F, 
    GeForce RTX 3060 12G GPU + 64GB memory + Intel Core i7-12700,
    GeForce RTX 2070 SUPER + 32GB memory + Intel Core i7-9700,
    GeForce RTX 2060 + 64GB memory + Intel Core i7-8700.
For experiments in highway, we use CPUs.

\section{More details about baselines and ablations}\label{appendix:baselines}

\paragraph{\BC{}.} 
For BC, we ran 1 seed for the replay buffer obtained at the end of training of \ours{}, therefore the performance is averaged over 5 different label buffers.
In Atari, we select the best result during training with respect to during training the accuracy to reach $\AccTarget=0.999$.
In highway, since we have multiple performance metrics and it's not easy to measure which one is the best, we present the performance when it reaches our default $\AccTarget=0.98$.

\paragraph{\Rainbow{}.}
We compare \ours{} with Rainbow in \Cref{tab:compareDQNs}.
See \Cref{tab:hyperRainbow} for hyper-parameters of \Rainbow{}. We use the same setting with a data-efficient version of Rainbow, except for the step $\StepN$ and the training timesteps (see \Cref{tab:HypersDiffEnv}).
% \begin{adjustbox}{width=\textwidth,center,keepaspectratio,caption={Compare the different component between \ours{} and Rainbow.},float=table}
    \begin{table}[ht]
        \caption{Compare the different component between \ours{} and Rainbow \citep{hessel2018Rainbow}.}
        \label{tab:compareDQNs}
        \centering
        \scalebox{0.9}{
            \begin{tabular}{llllllll}
            \toprule
            \multirow{2}{*}{Method}   & \multicolumn{7}{c}{Components}                                                                         \\
            \cline{2-8}
                                      & 1-step    & N-step   & Duel & Double Q & Replay              & Distrib & Exploration    \\
            \midrule
            \ours{} & Yes & Yes & Yes  & No & Uniform & Yes & Feedback guided \\
            % PureRL                    & Yes          & Yes          & Yes  & No       & uniform over recent & Yes             & no exploration \\
            % Q-learning (uni)          & Yes          & Yes          & Yes  & Yes      & uniform             & Yes             & epsilon-greedy \\
            % Q-learning (prior)        & Yes          & Yes          & Yes  & Yes      & prior               & Yes             & epsilon-greedy \\
            Rainbow & Yes & Yes & Yes & Yes & Prioritized  & Yes & Noisy-net \\
            \bottomrule
            % \caption{Compare detailed configuration with Deep Q-learning methods}
            \end{tabular}
        }
    \end{table}
% \end{adjustbox}

\paragraph{Compare design choices with baselines and ablations.}
DQfD \citep{hester2018DQfD} is a method that learns from human demonstrations, 
and PVP \citep{peng2023PVP} learns from human interventional control.
Both of the two methods are designed by augmenting the standard RL loss with their extra IL loss into normal RL methods like DQN \citep{mnih2015DQN} and TD3 \citep{fujimoto2018TD3}.
We compare the learning scheme and corresponding losses for each baselines and ablation settings in \Cref{tab:BaselineDesignChoice}, other hyper-parameters keep the same with \Cref{tab:BasicHyperparam} to have a fair comparison.
% \CMT{zj: Explain ablations since I cite it in the main text}
% \begin{adjustbox}{width=\textwidth,center,keepaspectratio,caption={Compare the design choices between \ours{} and different baselines.},float=table}
    \begin{table}[ht]
    \centering
    \caption{Compare the design choices between \ours{}, different baselines, and extra ablation groups. \zj{Backslashes mean there is no such phase.}}
    \label{tab:BaselineDesignChoice}
    \scalebox{0.86}{
        \begin{tabular}{l|l|cc}
        \toprule
        \multicolumn{2}{c|}{Methods} & \ours{} & \DQfD{} \\
        \midrule
        \multirow{2}{*}{Phases} & $\PhaseOne$ & $\LossMargE$ & $\backslash$  \\
                                        & $\PhaseTwo$ & $0.5\cdot\LossMargE+0.5\cdot\LossMargTGT+\LossOneTD+\LossNTD$ & $\LossMargE+\LossOneTD+\LossNTD$ \\
        \toprule
        \multicolumn{2}{c|}{Methods} & \PVPwR{} & \PVPwoR{} \\
        \midrule
        \multirow{2}{*}{Phases} & $\PhaseOne$ & $\backslash$ & $\backslash$ \\
                                        & $\PhaseTwo$ & $\LossPVP+\LossOneTD+\LossNTD$ & $\LossPVP+\LossOneTDwoR+\LossNTDwoR$ \\
        \toprule
        \multicolumn{2}{c|}{Methods} & DAgger &  \AblaNoFT{}\\
        \midrule
        \multirow{2}{*}{Phases} & $\PhaseOne$ & $\LossMargE$ & $\backslash$ \\
                                        & $\PhaseTwo$ & $\backslash$ & $0.5\cdot\LossMargE+0.5\cdot\LossMargTGT+\LossOneTD+\LossNTD$ \\
        \midrule
        \multicolumn{2}{c|}{Methods} &  \AblaNoTGT{} &\\
        \midrule
        \multirow{2}{*}{Phases} & $\PhaseOne$ & $\LossMargE$ &  \\
                                        & $\PhaseTwo$ & $\LossMargE+\LossOneTD+\LossNTD$ &  \\
        \bottomrule
        \end{tabular}
        }
    \end{table}

% \subsection{More discussions about related work}\label{appendix:discussion_related_work}
% \paragraph{RLIF}
% \citet{luo2024RLIF} \CMT{zj: add more contents here. Add the results I ran for RLIF.}

\zj{

\section{More Details about Experimental Settings}\label{appendix:environmental_setting}

\subsection{Configurations for evaluation}\label{appendix:EvaluationDetails}
We use 5 seeds for \ours{} and baselines for evaluation, and 3 seeds for ablations.
For all experiments we perform, we use 50 episodes in each evaluation.
When we ran experiments for \ours{} and baselines, we used 5 seeds for each setting.
For ablation studies, we use 3 seeds.
Although \ours{} and its' ablation \AblaNoFT{} involve 2 separate phase in each iteration, their performance are still evaluated at the end of each iteration, which is the $\phaseTwo$-phase.

\subsection{Highway}
Highway \citep{highway-env} is an environment 
    with state-based inputs in 35 dimensions and 5 available action choices to control the vehicle's speed and direction.
The input states contain information about both the controlled vehicles and the top 5 nearest vehicles' positions, speeds, and directions.
The basic goal in this environment is to drive a car on a straight road with multiple lanes within a limited time.
One episode is terminated when the time is used up, or a crash happens.

We use the configuration listed in \Cref{tab:highwayEnvConfig} in highway.
\begin{table}[ht]
\centering
\caption{Hyper-parameters for environment in highway.}
\label{tab:highwayEnvConfig}
% \vskip 0.1in
\scalebox{0.9}{
\centering
    \begin{tabular}{ll|c}
    \toprule
    \multicolumn{2}{c|}{Hyper-parameter}           & Value       \\
    \midrule
    \multirow{7}{*}{Env} & Available speed range  & [19,30]       \\
     & High speed range & $\ge$21    \\
     & Low speed range  & $<$21  \\
     & Lanes count   & 5           \\
     & Vehicles count & 40           \\
    & Time limit          & 50          \\
    & Policy frequency   & 1 \\
    \midrule
    \multirow{3}{*}{Observation} & Type  & Kinematics matrix     \\
     & Number of observed vehicles & 5        \\
     & Features for each vehicle & [presence, x, y, $v_x$, $v_y$, $\cos_{heading}$, $\sin_{heading}$]\\
     \midrule
     Action & Type & [Right, Left, Faster, Slower, IDLE]\\
    \bottomrule     
    \end{tabular}
    }
\end{table}

\subsection{Atari}\label{appendix:AtariEnvDetails}
In Atari, we use the signed raw reward from the environment as the proxy reward in our setting: 
    $\proxyrAtari=sign(\rawrAtari)$, where $\rawrAtari$ is the raw reward in Atari and $\proxyrAtari$ is the proxy reward.
The average cumulative $\rawrAtari$ of episodes serves as a performance metric.
    % \footnote{This train-evaluation setting is, in fact, the most standard one when evaluating RL algorithms in Atari.}.
% Except for \pong{} and \freeway{}, the other 8 games' $\proxyrAtari$ is not equal to 

Both $\rawrAtari$ and $\proxyrAtari$ have different levels of imperfection.
The imperfection of $\proxyrAtari$ mainly comes from two aspects:
    (1) Missing rewards, 
        e.g. in \seaquest{}, $\rawrAtari(LooseLives)=0$, but give a negative reward is better;
    (2) Numerical mismatch between $\proxyrAtari$ and $\rawrAtari$, 
        e.g., in \hero{}, $\rawrAtari(RescueMiners)=1K$, $\rawrAtari(ShootCritters)=50$ but $\proxyrAtari=1$ for both the two cases.
More information about the ground-truth $\rawrAtari$ can be checked in the official website\footnote{\href{https://gymnasium.farama.org/environments/atari/}{https://gymnasium.farama.org/environments/atari/}}.}
% \CMT{zj: maybe add more information about the env and their raw rewards, proxy rewards.}}

Our environment configurations follow standard requirements \cite{machado2018revisitALE}.
\Cref{tab:hyper_env_wrapper_atari} lists the related details.
\begin{table}[ht]
    \caption{Hyper-parameters for the Atari environment wrapper.}
    \label{tab:hyper_env_wrapper_atari}
    \centering
        \begin{tabular}{l|c}
        \toprule
        Hyper-parameter           & Value       \\
        \midrule
        Grey-scaling              & True       \\
        Observation down-sampling & (84, 84)    \\
        Frame stacked             & 4           \\
        Frame skipped             & 4           \\
        Action repetitions         & 4           \\
        Max start no ops          & 30          \\
        Reward clipping           & {[}-1, 1{]} \\
        Terminal on loss of life  & True        \\
        Max frames per episode    & 108K        \\
        \bottomrule     
        \end{tabular}
\end{table}

\subsection{Simulated labelers}\label{appendix:SimulatedExperts}
We use Rainbow to train \zj{labeler}s for all environments, with the general hyper-parameters shown in \Cref{tab:hyperRainbow} and concrete training timesteps in \Cref{tab:HypersDiffEnv}.
For highway, \zj{script labeler}s are trained with PRExp mentioned in \Cref{tab:HighwayReward1ChangeLane,tab:HighwayReward2RightLane}.
For Atari, \zj{script labeler}s are trained with the signed raw rewards.
Note that these script labelers are not necessarily perfect ones since their performance could be improved further with longer training.
% \CMT{zj: maybe add a table to compare our labelers' performance and Rainbow's}
% Note that in our experiments, we do not consider the ``optimality'' of the \zj{labeler}s because our goal is the performance alignment problem with the labeler.
% Although an \zj{labeler} achieve higher proxy reward do achieve better performance (like the experiments in Pong) (in fact, our simulated \zj{labeler}s have different levels of optimality). As a phenomenon observed in our experiments
\begin{table}[ht]
    \caption{Hyper-parameters for Rainbow.}
    \label{tab:hyperRainbow}
    \scalebox{0.84}{
    \centering
        \begin{tabular}{l|l}
        \toprule
        Hyper-parameter           & Value       \\
        \midrule
        Update period for target network  & per 2000 updates \\
        Atoms of distribution & 51        \\
        $\gamma$ & 0.99    \\
        Batch size             & 32           \\
        Optimizer             & Same with \Cref{tab:BasicHyperparam}          \\
        Max gradient norm & 10 \\
        Prioritized replay & exponent: 0.5, correction: 0.4$\rightarrow$1\\
        Noisy nets parameter & 0.1\\
        Warm-up steps & 1.6K\\
        Replay buffer size & 1M\\
        step $\StepN$ & Same with \Cref{tab:HypersDiffEnv} \\
        Sample steps to update & 1\\
        \midrule
        % \multirow{2}{*}{Netowrk-encoder} & Atari: channels [32, 64], kernel size [5, 5], strides [5,5], paddings [3, 1]\\
        \multirow{2}{*}{Netowrk-Atari} & CNN encoder: channels [32, 64], kernel size \& strides [5, 5], paddings [3, 1]\\
                                        & MLP policy: hidden size [128]\\
        \midrule
        \multirow{2}{*}{Netowrk-Highway} & MLP encoder: hidden size [128, 128]\\
                                         & MLP policy: hidden size [128]\\
        \bottomrule     
        \end{tabular}
        }
\end{table}

\section{More Details about User Study}\label{appendix:UserStudy}
\subsection{Evaluation Videos}\label{appendix:UserStudyVideos}
Evaluation videos for \oursHuman{} are sorted in \texttt{SupplementaryMaterials/Videos[Env]UserStudy}, which learns to perform more like human comparing with evaluation videos shown in \texttt{SupplementaryMaterials/Videos[Env]Oracles}.

\subsection{Procedure of User study}\label{appendix:UserStudyProcedure}
% \subsection{Ethics statement}
% We invite 2 participants to take part in our user study.
% They get paid, and they are free to stop the experiments anytime they want. Therefore, no discomfort happens.
% They only need to click the mouse on the interface (See \Cref{fig:UserInterface}) we designed; therefore no safety issue.
% \CMT{zj: I removed the Ethics statement part.}
No need to let humans play the game by themselves before providing feedback. 
Before letting the human labelers give feedback, we
    show them an instruction about the task they will participate (see \Cref{box:InstructPong,box:InstructHighway}),
    as well as an example video to let them become familiar with the environment dynamics (see \texttt{PATH/TO/SupplementaryMaterial]/Video[EnvName]Oracles-Examples}).
% \CMT{Add instruction for supplementary materials.}

\begin{figure}[ht]
    \begin{subfigure}{\textwidth}
      \centering
      \includegraphics[width=\linewidth]{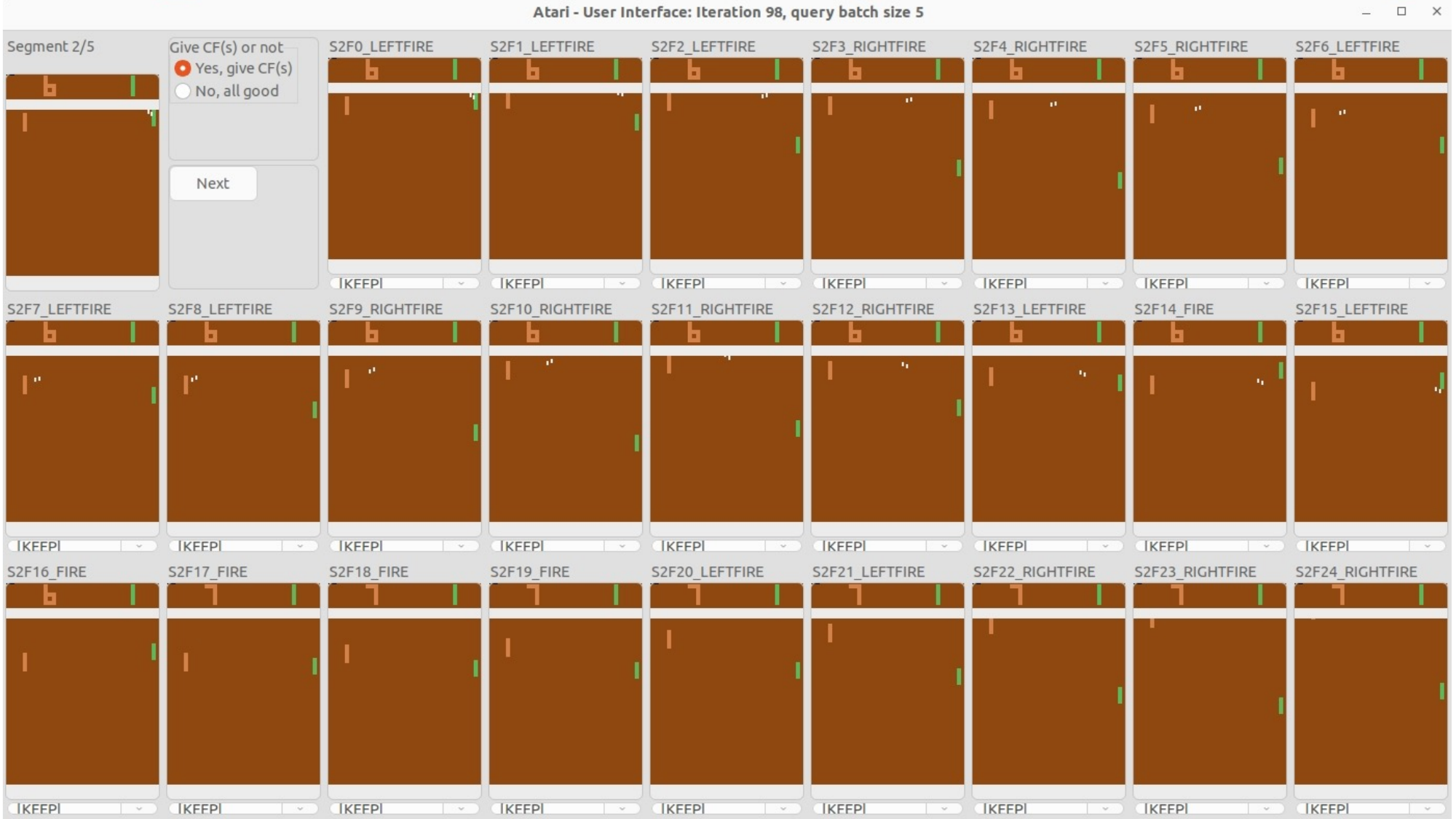}  
      \caption{User interface for \pong{}.}
      \label{fig:UserInterfacePong}
    \end{subfigure}
    \newline
    \begin{subfigure}{\textwidth}
      \centering
      \includegraphics[width=\linewidth]{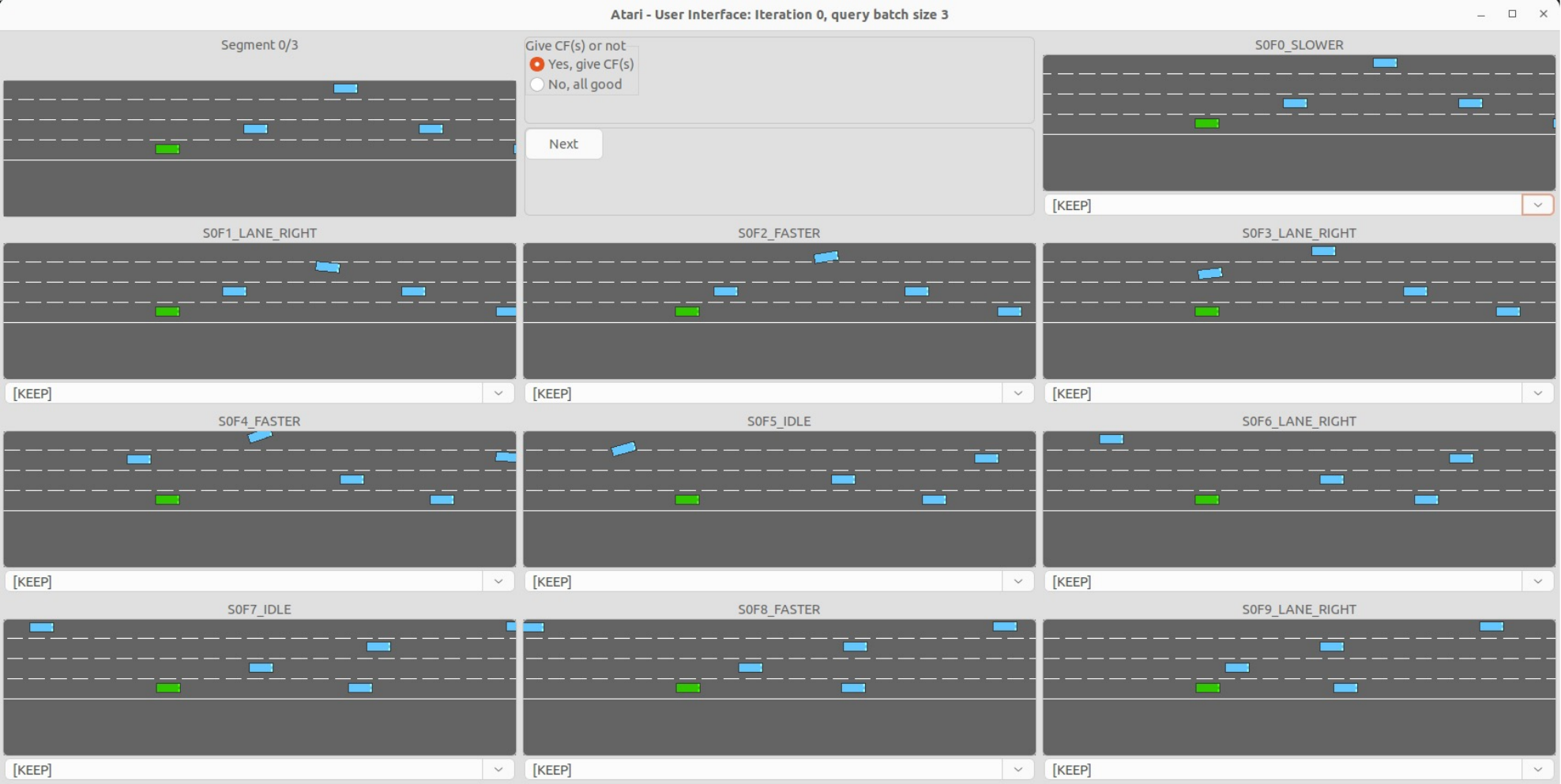}  
      \caption{User interface for highway.}
      \label{fig:UserInterfaceHighway}
    \end{subfigure}
\caption{User Interface. We put the video of that segment in the top left corner. If the labeler is satisfied with the whole segment's state-action pairs, they can choose to pass this segment with the radio box at the top near the video window. Otherwise the labeler will give corrective actions on other windows.}
\label{fig:UserInterface}
\end{figure}

We show our simple interface used in our user study in \Cref{fig:UserInterface}.
% \subsection{User Instructions}
% For the Atari environments, instructions are adapted from the official 
%     gymnasium \footnote{https://gymnasium.farama.org/} and 
%     Atari \footnote{
%     Pong: https://atariage.com/manual\_html\_page.php?SoftwareLabelID=587
%     % , Battlezone: https://atariage.com/manual\_html\_page.php?SoftwareID=859
%     } 
%     website, which contains the basic environmental setup, meaning of actions, and rewards.
\begin{table}[ht]
    \centering
    \caption{User instruction for \pong{}.}
    \label{box:InstructPong}
\begin{tcolorbox}[title=User Instructions - Pong] %tcolorbox
    \textbf{Description}: 
        \begin{itemize}
            \item You control the right paddle, you compete against the left paddle.
            \item You each try to keep deflecting the ball away from your goal and into your opponent’s goal.
        \end{itemize}
        
    \tcblower  % dashed line (separator)
    \textbf{Reward}: 
        \begin{itemize}
            \item A player scores +1 when the opponent hits the ball out of bounds or misses a hit, or -1 when you hits the ball out of bounds or misses a hit.
            \item The first player or team to score 21 points wins the game.
            % \item The last player or team to score always serves the ball.
            %         Serve by \verb|FIRE| after waiting at least one second after the point is made.
        \end{itemize}
    \tcbline
    \textbf{Actions}:
        \begin{table}[H]  % [H] need \usepackage{float}
            \centering
            \begin{tabular}{cccccc}
                \toprule
                Value & Meaning & Value & Meaning & Value & Meaning\\
                \midrule
                 0 & \verb|NOOP| & 1 & \verb|FIRE| & 2 & \verb|RIGHT| \\
                 3 & \verb|LEFT| & 4 & \verb|RIGHTFIRE| & 5 & \verb|LEFTFIRE|\\
                 \bottomrule
            \end{tabular}
        \end{table}
        \begin{itemize}
            \item \verb|NOOP|: random action
            \item \verb|LEFT|: move your paddle down
            \item \verb|RIGHT|: move your paddle up
            \item \verb|FIRE|: add some speed to the return ball or put sharper angles on your return hits when the ball contacts with your paddle
            \item Other actions are combined effects as described above.
        \end{itemize}
\end{tcolorbox}
\end{table}

\begin{table}[ht]
    \centering
    \caption{User instruction for highway.}
    \label{box:InstructHighway}
\begin{tcolorbox}[title=User Instructions - highway] %tcolorbox
    \textbf{Description}: 
        \begin{itemize}
            \item You control the green vehicle. There are other blue vehicles around.
            \item You need to drive at a speed as large as you can while avoiding a crash.
            \item You need to take over other vehicles if possible.
        \end{itemize}
        
    \tcblower  % dashed line (separator)
    \textbf{Reward}: 
        \begin{itemize}
            \item -1 for collision and 0 otherwise
        \end{itemize}
    \tcbline
    \textbf{Actions}:
        \begin{itemize}
            \item \verb|LANE_LEFT|: Change lane to the left, no effect if already in the leftmost lane.
            \item \verb|LANE_RIGHT|: Change lane to the right, no effect if already in the rightmost lane.
            \item \verb|IDLE|: No action, keep the current speed and heading direction.
            \item \verb|FASTER|: Faster.
            \item \verb|SLOWER|: Slower.
        \end{itemize}
\end{tcolorbox}
\end{table}

\clearpage

\end{document}